\documentclass{article}
\usepackage[preprint]{iclr2026_conference}

\usepackage{microtype}
\usepackage{graphicx}
\usepackage{subcaption}
\usepackage{booktabs}
\usepackage[hidelinks]{hyperref}
\usepackage{url}
\usepackage{float}
\usepackage{comment}
\usepackage{rotating}
\usepackage{caption}
\usepackage{xcolor}
\usepackage{colortbl}
\usepackage{tikz}
\usepackage{adjustbox}
\usepackage{multirow}
\usepackage{comment}
\usepackage{tabularx}
\usepackage{amssymb}
\usepackage[most]{tcolorbox}
\usepackage{lipsum}
\usepackage{pifont}
\tcbuselibrary{skins,breakable}

\usepackage{pgfplots}
\usepgfplotslibrary{groupplots}
\usetikzlibrary{patterns}
\pgfplotsset{compat=1.18}

\usepackage{makecell}

\hypersetup{colorlinks=true,linkcolor=blue,citecolor=blue,urlcolor=blue}

\title{CAMEL: An ECG Language Model \\ for Forecasting Cardiac Events}

\author{
    Neelay Velingker\thanks{Equal contribution and shared first authorship}, Alaia Solko-Breslin\footnotemark[1],
    Mayank Keoliya\footnotemark[1], Seewon Choi\footnotemark[1]\\
    \AND Jiayi Xin, Anika Marathe, Alireza Oraii,
    \AND Rajat Deo\thanks{Equal contribution and shared senior authorship}, Sameed Khatana\footnotemark[2], Rajeev Alur\footnotemark[2], Mayur Naik\footnotemark[2], Eric Wong\footnotemark[2] \vspace{1mm} \\
    University of Pennsylvania \\
    {\{\texttt{neelay,alaia,mkeoliya,seewon\}@seas.upenn.edu}}
}

\definecolor{mygreen}{HTML}{D5E8D4}
\definecolor{myred}{HTML}{F8CECC}
\definecolor{myorange}{HTML}{FFE0B8}
\definecolor{myblue}{HTML}{BBDCF7}
\definecolor{mypurple}{HTML}{E1D5E7}
\definecolor{myyellow}{HTML}{FFF2CC}

\definecolor{mydeepgreen}{HTML}{82B366}
\definecolor{mydeepred}{HTML}{B85450}
\definecolor{mydeeporange}{HTML}{F0A34A}
\definecolor{mydeepblue}{HTML}{5FA8E0}
\definecolor{mydeeppurple}{HTML}{9673A6}
\definecolor{mydeepyellow}{HTML}{D6B656}

\definecolor{promptbg}{RGB}{245,245,250}
\definecolor{responsebg}{RGB}{240,248,255}
\definecolor{bordercol}{RGB}{220,220,235}

\newtcolorbox{llmprompt}{
  enhanced,
  colback=promptbg,
  colframe=bordercol,
  arc=2mm,
  boxrule=0.5pt,
  leftrule=3pt,
  width=0.95\textwidth,
  enlarge left by=-1cm,
  enlarge right by=-1cm,
  title={\textbf{\textcolor{black}{Prompt}}},
  attach boxed title to top left={yshift=-2mm, xshift=4mm},
  boxed title style={colback=promptbg, colframe=bordercol, size=small, boxrule=0.5pt},
  breakable
}

\newtcolorbox{bigllmprompt}{
  enhanced,
  breakable,
  colback=promptbg,
  colframe=bordercol,
  coltext=black,       
  coltitle=black,      
  arc=2mm,
  boxrule=0.5pt,
  leftrule=3pt,
  width=\textwidth,
  left=4pt, right=4pt, boxsep=6pt,
  title={\textbf{System Prompt}},
  attach boxed title to top left={yshift=-2mm, xshift=4mm},
  boxed title style={
    colback=promptbg,
    colframe=bordercol,
    size=small
  }
}

\newtcolorbox{blueprints}{
  enhanced,
  breakable,
  colback=promptbg,
  colframe=bordercol,
  coltext=black,       
  coltitle=black,      
  arc=2mm,
  boxrule=0.5pt,
  leftrule=3pt,
  width=\textwidth,
  left=4pt, right=4pt, boxsep=6pt,
  title={\textbf{Blueprints}},
  attach boxed title to top left={yshift=-2mm, xshift=4mm},
  boxed title style={
    colback=promptbg,
    colframe=bordercol,
    size=small
  }
}

\newtcolorbox{multi-turn-conversation}{
  enhanced,
  breakable,
  colback=promptbg,
  colframe=bordercol,
  coltext=black,       
  coltitle=black,      
  arc=2mm,
  boxrule=0.5pt,
  leftrule=3pt,
  width=\textwidth,
  left=4pt, right=4pt, boxsep=6pt,
  title={\textbf{Conversation}},
  attach boxed title to top left={yshift=-2mm, xshift=4mm},
  boxed title style={
    colback=promptbg,
    colframe=bordercol,
    size=small
  }
}

\newtcolorbox{llmresponse}{
  enhanced,
  colback=responsebg,
  colframe=bordercol,
  arc=2mm,
  boxrule=0.5pt,
  leftrule=3pt,
  width=\textwidth+1cm,
  enlarge left by=-0.5cm,
  enlarge right by=-0.5cm,
  title={\textbf{\textcolor{black}{Response}}},
  attach boxed title to top left={yshift=-2mm, xshift=4mm},
  boxed title style={colback=responsebg, colframe=bordercol, size=small, boxrule=0.5pt},
  breakable
}

\tcbset{
  mybox/.style={
    enhanced,
    sharp corners,
    boxrule=0.4pt,
    colframe=black!50,
    colback=white,
    fonttitle=\bfseries,
    top=4pt,
    bottom=4pt,
    left=6pt,
    right=6pt,
    before skip=10pt,
    after skip=10pt
  }
}

\newcommand{\model}{CAMEL}
\newcommand{\penn}{Penn}
\newcommand{\cmark}{\ding{51}}
\newcommand{\xmark}{\ding{55}}

\begin{document}

\maketitle
\setcounter{footnote}{0}

\begin{abstract}
Electrocardiograms (ECG) are electrical recordings of the heart that are critical for diagnosing cardiovascular conditions.
ECG language models (ELMs) have recently emerged as a promising framework for ECG classification accompanied by report generation.
However, current models cannot \emph{forecast future cardiac events} despite the immense clinical value for planning earlier intervention.
To address this gap, we propose \textbf{\model{}}, the first ELM that is capable of inference over longer signal durations which enables its forecasting capability.
Our key insight is a specialized ECG encoder which enables cross-understanding of ECG signals with text. 
We train \model{} using established LLM training procedures, combining LoRA adaptation with a curriculum learning pipeline.
Our curriculum includes ECG classification, metrics calculations, and multi-turn conversations to elicit reasoning.
\model{} demonstrates strong zero-shot performance across 6 tasks and 9 datasets, including \textbf{ECGForecastBench}, a new benchmark that we introduce for forecasting arrhythmias. 
\model{} is on par with or surpasses ELMs and fully supervised baselines both in- and out-of-distribution, achieving SOTA results on ECGBench (+7.0\% absolute average gain) as well as ECGForecastBench (+12.4\% over fully supervised models and +21.1\% over zero-shot ELMs).
\footnote{Code and datasets are available at \href{https://github.com/CAMEL-ECG/CAMEL-inference} {github.com/CAMEL-ECG/CAMEL-inference} and \href{https://huggingface.co/CAMEL-ECG}{huggingface.co/CAMEL-ECG}.}
\end{abstract}

\section{Introduction}
\label{sec:intro}

Electrocardiograms (ECG) are multi-dimensional recordings of the heart's electrical activity and serve as a primary tool for diagnosing and triaging conditions such as heart attacks, arrhythmias, and other cardiac abnormalities \citep{berkaya2019, stefano1999ecg}.
From symbolic algorithms and statistical models to CNNs, automated ECG classification has moved from extensive academic study to widespread deployment in both ambulatory and in-hospital environments, as seen in systems like GE’s Marquette 12SL \citep{ge2019marquette}. More recently, foundation models have emerged for jointly processing ECG and text, which we call \emph{ECG Language Models (ELMs)}.
ELMs combine ECG representation learning with natural language generation to produce interpretable classifications and reports \citep{pulse,liu2024merl,wang2025melp,gem-empowering-mllm}.
Despite their promise, existing ELMs only target classification and do not predict a patient's future state, thus offering limited support for early intervention.

Forecasting cardiac events from ECG signals is a key challenge for AI in cardiac care.
Unlike classification, forecasting requires detecting subtle, prognostic patterns in ECGs to anticipate future adverse events.
Such early warning of cardiac events, such as ventricular tachycardia, could allow clinicians to intervene to improve patient outcomes \citep{Pollack2016-pediatric-risk-mortality-score,Soar2021-adult-advanced-life-support}.
While classical ML models and CNNs have been applied to this task \citep{jhu-picu,forecasting-imminent-afib}, they rely on fully supervised training for fixed-length inputs and offer interpretability only through post-hoc explanations, limiting their ability to generalize across tasks and clinical contexts.
In contrast, ELMs contain an LLM backbone trained on clinical knowledge, allowing them to generalize across tasks and generate natural language explanations along with their predictions.

To meet this challenge, we propose \model{} (\underline{C}ardiac \underline{A}utoregressive \underline{M}odel for \underline{E}CG \underline{L}anguage‑Modeling), the first general-purpose ELM designed to support long temporal context windows of ECG signals.
While existing benchmarks are largely restricted to the classification of 10-second snippets, we introduce ECGForecastBench, a new benchmark for predicting future arrhythmias from baseline normal sinus rhythms as input.
Our model generates forecasting reports by leveraging ECG signal statistics with established clinical associations \citep{afib-hrv-forecast}.
These statistics provide physiologically grounded explanations for the risk of a future cardiac event.

\begin{figure*}[t] 
    \centering
    \includegraphics[width=0.95\textwidth]{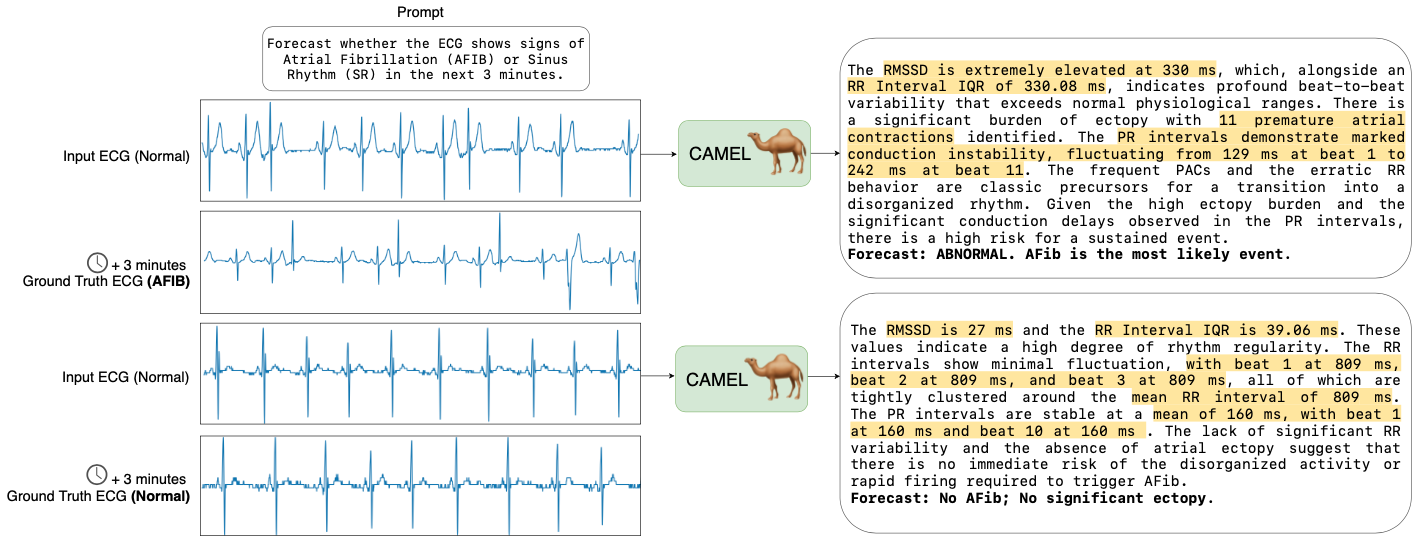}
    \caption{Example of \model's forecasting capability.  In the top example, \model{} takes as input normal sinus rhythm ECG at time $T$ and correctly forecasts AFIB at $T+3$ minutes by reasoning over the RMSSD, RR-interval, and PAC count (reasoning highlighted).
    In the bottom example, \model{} correctly predicts a normal outcome based on accurately extracted statistics.}
    \label{fig:overview}
\end{figure*}

Like prior ELMs, \model{} builds on a pre-trained large language model backbone, namely MedGemma-4B \citep{medgemma}, to support reasoning and natural language generation.
The core insight that allows \model{} to reason over long temporal contexts of ECG signals is how the integration of signal embeddings with text embeddings operates at the token level.
By encoding each one-second segment of each lead in an ECG as an individual token, \model{} can interleave multiple signal sequences of any duration with textual prompts.
This design supports flexibility in both input length and lead configuration, enabling \model{} to reason over long ECG contexts and variable, potentially incomplete sets of leads common in real-world settings.
This is in contrast to prior ELMs, whose contexts are generally restricted to 10-second, 12-lead ECGs (Table \ref{tab:ecg_models}).

To train \model{}, we introduce a 5-stage curriculum that gradually builds the model's reasoning and forecasting capabilities.
Training starts with an autoencoder stage to learn robust ECG representations.
Subsequent stages teach the model multiple-choice and short-answer tasks, understanding of ECG statistics, multi-turn conversational reasoning, and finally the generation of forecasting reports.
This curriculum enables \model{} to compute ECG statistics from long contexts and use them as evidence for clinically grounded forecasts (Fig. \ref{fig:overview}).

In summary, the main contributions of this paper are as follows.
First, we introduce the architecture for \model, which enables its unique capability to reason over long-duration ECG signals and identify predictive markers of future events.
Next, we present a large-scale data generation pipeline that supports curriculum learning for ECG comprehension, including the development of a novel benchmark, ECGForecastBench, for evaluating forecasting of future adverse cardiac events.
We then present a staged training algorithm that progressively builds \model's ECG grounding, reasoning, and forecasting capabilities.
Finally, we demonstrate that \model\ achieves strong zero-shot performance, matching or surpassing fully supervised in- and out-of-distribution baselines.

\section{\model{} Architecture}
\label{sec:model}
In this section, we describe \model's architecture and detail the characteristics that enable its unique capabilities.

\newcommand{\betweenfs}{\fontsize{8.0}{10}\selectfont}
\setlength{\tabcolsep}{5pt}
\begin{table*}[t]
\centering
\caption{Comparison of ECG foundation models across capabilities. The duration field indicates the ECG lengths supported during training and evaluation for each model. \model{} is the first ELM to support forecasting, multi-ECG inputs, and long-context signals.}
\footnotesize
\label{tab:ecg_models}
\begin{tabular}{l|ccc|ccc}
\toprule
\textbf{} & \multicolumn{3}{c|}{\textbf{Tasks}} & \multicolumn{3}{c}{\textbf{Inputs}} \\
\midrule
\textbf{Model} & 
\makecell{\textbf{Input} \\ \textbf{Classification}} & 
\makecell{\textbf{Report} \\ \textbf{Generation}} & 
\makecell{\textbf{Forecasting}} &
\textbf{Modality} &
\textbf{\# ECGs} & 
\textbf{Duration (s)} \\
\midrule
ECGFounder \citeyearpar{li2025ecgfounder} & \cmark & \xmark & \xmark & waveform & single & 10 \\
ECG-FM  \citeyearpar{mckeen2025ecgfm}     & \cmark & \xmark & \xmark & waveform & single & 10 \\
MERL    \citeyearpar{liu2024merl}         & \cmark & \xmark & \xmark & waveform & single & 10 \\
ECG-Byte \citeyearpar{han2025ecgbyte}     & \cmark & \xmark & \xmark & waveform & single & 10 \\ 
MELP   \citeyearpar{wang2025melp}         & \cmark & \cmark & \xmark & waveform & single & 10 \\
PULSE \citeyearpar{pulse}                & \cmark & \cmark & \xmark & image, text & multi & 10 \\
GEM \citeyearpar{gem-empowering-mllm}     & \cmark & \cmark & \xmark & image, waveform, text & multi & 10 \\ 
\midrule 
\rowcolor{gray!20} \textbf{\model ~(Ours)} & \cmark & \cmark & \cmark & waveform, text & multi & 0--10K \\
\bottomrule
\end{tabular}
\end{table*}

\begin{figure*}[t]
    \centering
    \includegraphics[width=0.95\textwidth]{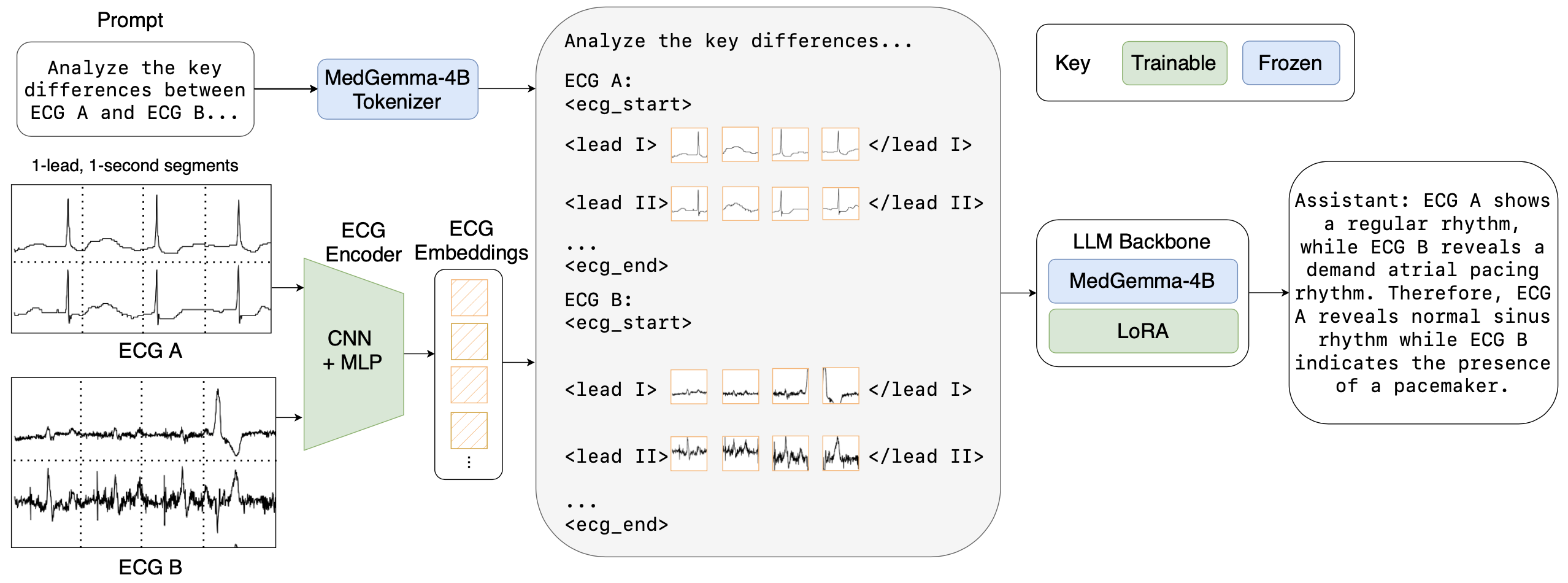}
    \caption{An overview of the \model{} architecture. 1-second, single-lead ECG segments from two patients are encoded and combined with text token embeddings. The resulting sequence is processed by an LLM backbone (MedGemma-4B with LoRA adapters) to generate a clinical report. Fixed models are shown in blue, and trainable models are shown in green.}
    \label{fig:architecture}
\end{figure*}

As Figure \ref{fig:architecture}, illustrates, \model~ consists of two main components:
1) a trainable tokenizing encoder that maps ECG segments into vector representations, and
2) a LoRA-adapted LLM that processes the aligned ECG embeddings along with text.
Specifically, \model{} is built upon MedGemma-4B \citep{medgemma}, leveraging its medical language understanding.
Together, these jointly trained components produce an end-to-end late-fusion pipeline for ECG-centric multi-turn conversation.

\paragraph{ECG encoder.} We consider a single-lead ECG signal sampled at $N$~Hz. A 1-second ECG segment is represented as $\boldsymbol{x}_{ecg} \in \mathbb{R}^N$.
To encode the raw waveform, we use a 3-layer CNN $f_\theta: \mathbb{R}^{N} \rightarrow \mathbb{R}^{d}$ which maps each 1-second segment into a latent representation:
\begin{equation*}
    \boldsymbol{z}_{ecg} \in \mathbb{R}^d = f_{\theta}(\boldsymbol{x}_{ecg}).
\end{equation*}
In our implementation, we use $d = 64$, and all signals are resampled to the same frequency $N = 256$.
This design yields a compact, continuous representation of ECG segments, preserving clinically meaningful features while requiring relatively few tokens to represent long signals.

To integrate ECG representations with the LLM, ECG embeddings must align with the LLM's hidden token space with dimension $h$. Therefore, we apply a linear projection layer to the ECG embeddings, implemented as a single linear layer $SLP_\phi : \mathbb{R}^{d} \rightarrow \mathbb{R}^h$
\begin{equation*}
    \boldsymbol{e}_{ecg} \in \mathbb{R}^h = SLP_\phi( \boldsymbol{z}_{ecg}).
\end{equation*}

\paragraph{Integrating ECG and text.} Given an ECG signal of $T$ seconds with $L$ leads, we encode each lead $\ell \in \{1, \dots, L\}$ as a sequence of 1-second embeddings: 
\begin{equation*}
\boldsymbol{H}_{ecg}^{(\ell)}
=
\big[
\boldsymbol{e}_{ecg}^{(\ell,1)},
\ldots,
\boldsymbol{e}_{ecg}^{(\ell,T)}
\big]
\in \mathbb{R}^{T \times h}.
\end{equation*}
To explicitly encode structure, we introduce lead start and end special tokens for each input lead, between which the encoded segments are placed:
\begin{equation*}
\tilde{\boldsymbol{H}}_{ecg}^{(\ell)}
=
\text{Concat}\!\left(
\boldsymbol{e}_{\langle \text{\texttt{lead}}_{\ell} \rangle},
\boldsymbol{H}_{ecg}^{(\ell)},
\boldsymbol{e}_{\langle \text{\texttt{/lead}}_{\ell} \rangle}
\right).
\end{equation*}

With this, we are ready to integrate the ECG with text. Given input text $\boldsymbol{x}_{txt}$, MedGemma's tokenizer maps each text token to an embedding in the same hidden space. 
Without loss of generality, \model{} combines these embeddings with ECG start and end special tokens to form a unified sequence suitable for LLM input,
\begin{equation*}
\boldsymbol{H} =
\text{Concat}\Big(
\text{Embed}(\boldsymbol{x}_{txt}),
\boldsymbol{e}_{\langle \texttt{ecg} \rangle},
\bigoplus_{\ell=1}^{L} \tilde{\boldsymbol{H}}_{ecg}^{(\ell)},
\boldsymbol{e}_{\langle \texttt{/ecg} \rangle}
\Big),
\end{equation*}
where $\bigoplus$ denotes concatenation.
Note that text may also be placed between ECGs or at the end of the sequence.
Furthermore, lead-specific special tokens may appear multiple times, allowing users to provide more than one ECG in a conversation, which can be placed anywhere in a user turn.
This design allows for comparative reasoning across ECGs, supporting richer ECG comprehension.

\textbf{Lead-aware attention masking.}
Unlike text, multi-lead ECG signals are not strictly causal when flattened. 
Specifically, all leads at a given time $t$ measure simultaneous projections of the same underlying cardiac electrical activity.
To leverage this insight,  and enable cross-lead learning, \model\ uses a custom attention mask that allows the tokens of 1-second ECG segments at the same position (time $t$) to attend to each other bidirectionally, regardless of prompt lead order.
Additionally, ECG embeddings can attend to their corresponding lead boundary markers and all preceding special tokens, while lead boundary markers can attend to all ECG embeddings within their corresponding leads.
This design allows the model to learn inter- and intra-lead relationships while preserving the sequential dependencies required for autoregressive text generation.

\section{Training} \label{sec:training}
In this section, we describe the 5-stage training pipeline we use to learn ECG representations and translate this to general ECG reasoning and forecasting.

\paragraph{Data preprocessing.}\label{par:preprocessing}
Prior to training, we perform minimal preprocessing to preserve the information in raw data as much as possible while allowing the model to generalize across diverse clinical settings. 
We apply 50/60 and 0.3Hz band-pass filters and resample each ECG to $256$ Hz.
We exclude data with persistent ($>$ 5 seconds) zero or NaN values and replace any remaining NaN values with zero. 

\paragraph{Stage 1: Autoencoder training.} In the first stage, we use self-supervised learning to train the the CNN-based encoder (i.e., the mapping from raw ECG to $\boldsymbol{z}_{ecg}$), which will later be used as part of the full CNN+SLP encoder.
Given 1-second, single-lead ECG segments, the encoder-decoder pair is optimized with MSE loss to reconstruct the input signal.
The training dataset includes over 1 billion ECG segments collected from 13 public datasets (Appendix \ref{appendix-experimental-set-up}).

For subsequent training stages, we connect the CNN-based encoder trained in stage 1 to the LLM via a linear layer, allowing ECG embeddings to be mixed in with text and special token embeddings.
We train using ECG-text instruction tuning using a masked autoregressive cross-entropy loss.
Unless otherwise specified, we apply lead-wise shuffling and random lead dropping during training to improve robustness to varying input leads.
See Appendix \ref{appendix-experimental-set-up} for further training details.

\paragraph{Stage 2: Multiple choice and short answer.}
In this next stage, we begin curriculum learning with simpler instructional formats and less challenging ECG-related content than latter stages.
Specifically, Stage 2 is restricted to short-answer and multiple-choice questions, primarily targeting high-level ECG classification tasks.
The data for Stage 2 are comprised of 20 million samples derived from the Harvard-Emory dataset \citep{harvard-emory-ecg-database}.

\paragraph{Stage 3: Adding statistics questions.}
In the third stage, curriculum training incorporates question formats that require fine-grained understanding of ECG statistics.
Specifically, we leverage the ECGDeli library \citep{ecgdeli} to compute lead-specific and global statistics for each sample of the datasets from Stage 2.
We then create short-answer questions based on these statistics, including statistics computation such as extracting T-wave amplitude, step-by-step reasoning e.g to extract beat-wise R-R interval and compute heart rate, and multi-ECG comparisons to compare e.g. longer QRS durations.
These questions teach clinically meaningful statistics that underlie the diagnoses introduced in Stage 2.
Further details about the Stage 3 data are in Appendix \ref{appendix-stage3}.

\paragraph{Stage 4: Multi-turn conversations.}
In Stage 4, we unify the classification and statistical reasoning capabilities by training \model{} with multi-turn conversations with 1 or 2 ECGs. We build on the GEM framework, which first introduced the generation of ECG reports grounded in ECG statistics. \citep{gem-empowering-mllm}.
Using the GE Marquette 12SL manual \citep{ge2019marquette}, we identify clinically important global and lead-specific statistics associated with each diagnosis, which guides the selection of statistics for conversations.
To ensure there is substantial diversity, each conversation randomly samples a task type, audience, response format, and dialogue arc (detailed in Appendix \ref{appendix-stage4}).
We pass these criteria, along with the selected statistics and diagnosis, to \texttt{gemini-3-flash-preview} to generate the dialogues.
Further details about the Stage 4 data are in Appendix \ref{appendix-stage4}.
We refer to the model trained up to Stage 4 as \model{}-Base.

\paragraph{Stage 5: Forecasting.}
In the final stage, we train the model to produce evidence-based narratives that reason about future cardiac rhythm transitions over specified time horizons.
To construct a forecasting dataset, we use the Icentia11k dataset containing single-lead tracings, spanning multiple days per patient, with beats labeled as normal, atrial fibrillation (AFIB), or atrial flutter (AFL) \citep{icentia11k}.
We compile literature-grounded rubrics covering known electrophysiologic substrates, triggers, and temporal risk factors for atrial arrhythmias.
These features are verified with cardiologists to ensure clinical plausibility.
We generate samples across various input lengths, forecast time horizons, and label transitions, e.g., normal $\rightarrow$ normal or normal $\rightarrow $ AFIB.
Using the validated features, we generate supervised forecasting examples with \texttt{gemini-3-flash-preview} consisting of a predicted transition label and a structured clinical explanation that links observed temporal trends to future risk (Fig. \ref{fig:overview}). 
Further details about the Stage 5 data are in Appendix \ref{appendix-stage5}.
We refer to the model trained up to Stage 5 as \model{}-Forecast.

\section{Evaluation}
\label{sec:evaluation}

In this section, we evaluate \model{} against various baselines on downstream tasks such as forecasting, classification, and report generation.

\subsection{Experimental Setup}
We use the MedGemma-4B-instruct model \citep{medgemma} as the pretrained LLM backbone. 
The ECG encoder is comprised of a 3-layer CNN which projects each 256 Hz 1-lead, 1-second ECG segment to a $d=64$ dimension latent space, which is then passed through an SLP to the LLM's latent token space ($h=2560$). 
All training is done on 32 B200 NVIDIA GPUs.
We fit the LLM backbone with LoRA adapters, each with a rank of $r=128$.
For further compute timing details and hyperparameters, see Appendix \ref{appendix:hyperparam}.
Unless otherwise specified, we evaluate using \model{}-Base on each task.
Given \model{}'s relatively small parameter size, fine-tuning on the forecasting dataset prioritizes task-specific performance and results in a modest tradeoff in open-ended conversational ability, rather than a loss of core reasoning ability.

\subsection{Downstream Tasks \& Baselines}

\paragraph{Forecasting.} 
We evaluate forecasting performance on the Icentia11k \citep{icentia11k} dataset. Icentia11k contains single-lead ECG recordings with NORM, AFIB, and AFL beat-level annotations. 
We report the F1 score for varying forecasting horizons (time-to-event) and input ECG signal duration.
For this task, we use the \model{}-Forecast model.

\paragraph{Classification.} 
We evaluate on classifying diagnostic codes using seven datasets: PTB-XL \citep{ptbxl}, CSN \citep{champman}, CODE-15\% \citep{code15}, CPSC-2018 \citep{cpsc2018}, HEEDB \citep{harvard-emory-ecg-database}, Icentia \citep{icentia11k}, and \penn.
\penn{} is a privately sourced ECG dataset from the University of Pennsylvania hospital system, containing {1K} 8-lead ECG recordings from ICU units with NORM, Ventricular Fibrillation (VT), and Ventricular Tachycardia (VF) labels.
See Appendix \ref{appendix:classification-setup} for more details.
We report F1 score for zero-shot evaluation and AUROC for linear probing results.
Like other baselines, we finetune \model{}-Base on ECGInstruct \citep{pulse} for this task.

\paragraph{Report generation.}
We evaluate report generation and conversation performance on PTB-XL \citep{ptbxl} and MIMIC-IV-ECG \citep{mimic-iv-ecg}. 
We utilize GPT-5 to score the responses on diagnostic accuracy, analysis completeness, and relevance based on the predefined criteria used for GEM \cite{gem-empowering-mllm}.
We also report traditional NLP metrics: BLEU-1, BLEU-4, METEOR, Rouge, and BERT-F1.
Like other baselines, we finetune \model{}-Base on ECGInstruct \citep{pulse} for this task.

\paragraph{Question answering.}
We use multiple-choice questions from ECGBench \citep{pulse}, sourced from the G12 \citep{challenge2020} and CSN \citep{champman} datasets and ECG-QA \cite{ecg-qa}, sourced from PTB-XL and MIMIC-IV-ECG. 
ECG-QA comprises single-ECG questions for general interpretation and multi-ECG questions for comparative analysis of two ECGs. 
We use accuracy and hamming score. 

\paragraph{Multi-turn conversations.}
This benchmark comprises two multi-turn task types curated from PTB-XL reports.
The first is a two-turn task that requires interpreting a single ECG and subsequently answering questions about it. 
The second is a three-turn task that requires interpreting two ECG recordings and performing comparative analysis.
Following PULSE \citep{pulse}, we use GPT-5 to score the conversation based on diagnostic accuracy, analysis completeness, and instruction adherence.

\paragraph{Grounding.}
We create a benchmark of 2K ECG statistics (RR interval, HR, QRS duration, QTc interval, PR interval, QRS amplitude) from CPSC-2018 samples.
We report RMSE between the ground truth values computed using ECGDeli \citep{ecgdeli} and the predicted values.

\paragraph{Baselines.}
For baselines, we use MELP \citep{wang2025melp}, MERL \citep{liu2024merl}, PULSE \citep{pulse}, and GEM \citep{gem-empowering-mllm}. MELP and MERL are capable of zero-shot classification of ECG waveforms. Both take a list of classes and ECG as input and return the most likely class based on the cosine similarity. PULSE and GEM are multi-modal LLMs finetuned for ECG interpretation. PULSE uses ECG images as inputs whereas GEM uses both images and waveforms. 
We use all four baselines for classification tasks, but exclude MERL and MELP on others where a prompt has to be supplied. On the forecasting task we  also evaluate on GPT-5.2, augmented with a Code Interpreter, and access to the raw ECG stored as a CSV file.

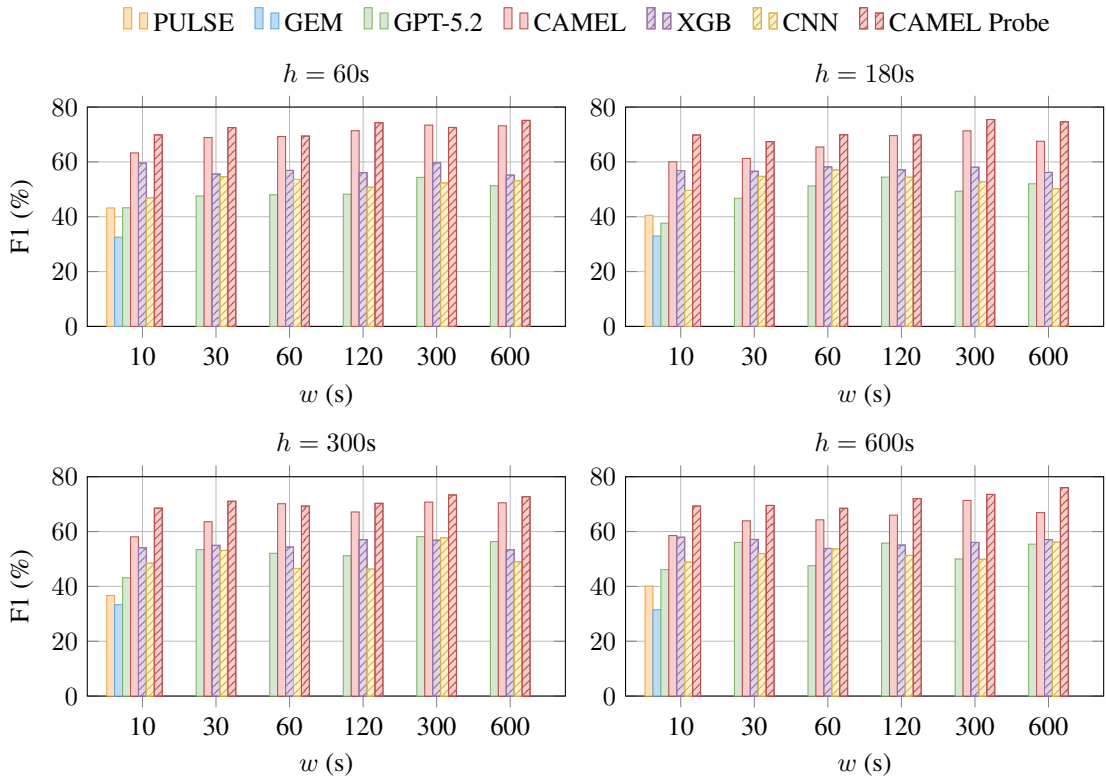
\begin{figure*}[t]
\centering
\begin{tikzpicture}
\centering
\pgfplotsset{
    pulse/.style={mydeeporange, fill=myorange},
    gem/.style={mydeepblue, fill=myblue},
    camel/.style={mydeepred, fill=myred},
    probe/.style={mydeepred, fill=myred, postaction={
    pattern=north east lines,
    pattern color=mydeepred
     }},
    gpt/.style={mydeepgreen, fill=mygreen},
    xgb/.style={mydeeppurple, fill=mypurple, postaction={
    pattern=north east lines,
    pattern color=mydeeppurple
     }},
    cnn/.style={mydeepyellow, fill=myyellow, postaction={
    pattern=north east lines,
    pattern color=mydeepyellow
     }},
}

\begin{groupplot}[
group style={
    group size=2 by 2,
    horizontal sep=0.8cm,
    vertical sep=2cm
},
ybar,
/pgf/bar width=3pt,
width=0.5\textwidth,
height=4.5cm,
symbolic x coords={10,30,60,120,300,600},
xtick={10,30,60,120,300,600},
enlarge x limits=0.15,
grid=major,
xlabel={$w$ (s)},
ymin=0, ymax=80,
ylabel near ticks,
legend style={
    at={(1.05,1.28)},
    anchor=south,
    legend columns=-1,
    draw=none,
    /tikz/every even column/.append style={column sep=0.2cm}
}
]

\nextgroupplot[title={$h=60$s}, ylabel={F1 (\%)}]
\addplot[pulse, bar shift=-12pt] coordinates {(10,43.21)};
\addplot[gem, bar shift=-9pt]coordinates {(10,32.54)};
\addplot[gpt, bar shift=-6pt] coordinates {(10,43.26) (30,47.54) (60,48.00)(120,48.24) (300,54.36) (600,51.28)};
\addplot[camel, bar shift=-3pt]coordinates { (10,63.28) (30,68.88) (60,69.23)(120,71.37) (300,73.42) (600,73.18)};
\addplot[xgb, bar shift=0pt] coordinates {(10,59.55) (30,55.58) (60,56.89) (120,56.09) (300,59.64) (600,55.20)};
\addplot[cnn, bar shift=3pt] coordinates {(10,46.89) (30,54.62) (60,53.59) (120,50.79) (300,52.35) (600,53.10)};
\addplot[probe, bar shift=6pt]coordinates {(10,69.83) (30,72.52) (60,69.43) (120,74.30) (300,72.56) (600,75.14)};
\legend{PULSE, GEM, GPT-5.2, \model{}, XGB, CNN, \model{} Probe}

\nextgroupplot[title={$h=180$s}]
\addplot[pulse, bar shift=-12pt]coordinates {(10,40.50)};
\addplot[gem, bar shift=-9pt] coordinates {(10,32.95)};
\addplot[gpt, bar shift=-6pt] coordinates {(10,37.67) (30,46.71) (60,51.22)(120,54.43) (300,49.29) (600,52.01)};
\addplot[camel, bar shift=-3pt]
coordinates {(10,60.04) (30,61.27) (60,65.44) (120,69.59) (300,71.34) (600,67.53)};
\addplot[xgb, bar shift=0pt] coordinates {(10,56.85) (30,56.60) (60,58.18) (120,57.09) (300,58.07) (600,56.17)};
\addplot[cnn, bar shift=3pt] coordinates {(10,49.59) (30,54.72) (60,57.08) (120,54.40) (300,52.76) (600,50.21)};
\addplot[probe, bar shift=6pt]coordinates { (10,69.81) (30,67.38) (60,69.91)(120,69.85) (300,75.46) (600,74.64)};

\nextgroupplot[title={$h=300$s}, ylabel={F1 (\%)}]
\addplot[pulse, bar shift=-12pt] coordinates {(10,36.67)};
\addplot[gem, bar shift=-9pt] coordinates {(10,33.33)};
\addplot[gpt, bar shift=-6pt] coordinates {(10,43.14) (30,53.42) (60,52.07) (120,51.19) (300,58.16) (600,56.37)};
\addplot[camel, bar shift=-3pt] coordinates {(10,58.07) (30,63.58) (60,70.15)(120,67.15) (300,70.73) (600,70.48)};
\addplot[xgb, bar shift=0pt] coordinates {(10,54.08) (30,54.97) (60,54.35) (120,57.05) (300,56.86) (600,53.37)};
\addplot[cnn, bar shift=3pt] coordinates { (10,48.47) (30,53.13) (60,46.54) (120,46.32) (300,57.72) (600,48.99)};
\addplot[probe, bar shift=6pt] coordinates {(10,68.54) (30,71.05) (60,69.31)(120,70.30) (300,73.36) (600,72.70)};

\nextgroupplot[title={$h=600$s}]
\addplot[pulse, bar shift=-12pt]coordinates {(10,40.12)};
\addplot[gem, bar shift=-9pt]coordinates {(10,31.49)};
\addplot[gpt, bar shift=-6pt] coordinates { (10,46.11) (30,56.04) (60,47.55) (120,55.80) (300,50.01) (600,55.41)};
\addplot[camel, bar shift=-3pt] coordinates { (10,58.54) (30,63.91) (60,64.28) (120,65.99) (300,71.37) (600,66.90)};
\addplot[xgb, bar shift=0pt] coordinates { (10,57.93) (30,57.12) (60,53.84) (120,55.06) (300,56.03) (600,57.07)};
\addplot[cnn, bar shift=3pt] coordinates {(10,48.85) (30,51.96) (60,53.69) (120,51.32) (300,49.90) (600,56.19)};
\addplot[probe, bar shift=6pt] coordinates {(10,69.30) (30,69.53) (60,68.49)(120,72.04) (300,73.54) (600,76.02)};

\end{groupplot}
\end{tikzpicture}
\caption{Forecasting performance (Macro-F1) in predicting AFib, AFlutter or Sinus Rhythm across input window $w $ $(s)$ and horizon $h$ $(s)$. We report zero-shot results from GPT-5.2 (with Code Interpreter and high effort), PULSE, GEM, and CAMEL, supervised training results for XGB and CNN, and linear probing results for CAMEL Probe. PULSE and GEM only support 10-second ECG inputs ($w = 10$). CAMEL outperforms all baselines, with increased performance with higher input windows, highlighting the importance of supporting longer ECG recordings. }
\end{figure*}

\subsection{Forecasting}
We study the effectiveness of \model{} on forecasting adverse cardiac events. 
We compare the performance as we vary the forecasting horizon $h \in \{1, 3, 5, 10\}$ minutes and ECG signal duration $l \in \{10, 30, 60, 120, 300, 600\}$ seconds in Table \ref{tab:appendix-forecasting}.
Note that unlike \model{}, baseline ELMs do not support ECG inputs longer than 10 seconds.

The results demonstrate that \model{} (evaluated on a subset of the test data), outperforms all baselines, including GPT-5.2 by over 10\%.
Across all models and datasets, we observe that the model performance drops as we increase the forecasting horizon.
However, we observe a notable increase in F1 score with longer ECG signals.
For 10 minutes inputs, we are providing 40 times more information compared to the standard 10 second ECGs, and the temporal variability in signal aids forecasting.

\subsection{Existing Benchmarks}

We compare zero-shot performance of \model{} against baseline models on previously studied tasks: classification, report generation, question answering, and grounded ECG understanding.

\paragraph{Classification.}
We present zero-shot classification results using 10-second, full-lead inputs in Tables~\ref{tab:comprehensive_eval} and \ref{tab:classification-full}. For Icentia11k and \penn{} datasets, we provide 1 and 8 leads respectively, and 12 leads for all other datasets.
We additionally evaluate reduced-lead settings to mimic scenarios where fewer leads are available, with results reported in Table~\ref{tab:classification-fixed}.
For Icentia11k AFIB classification, we vary the input duration $l \in \{10, 30, 60, 120, 300, 600\}$ seconds to assess the effect of temporal context (Table \ref{tab:appendix-length}).

In the single-lead telemetry analysis, the sensitivity for classifying AFIB increased as the recording duration increased from 10 to 60 seconds.
After 60 seconds, there were only modest improvements in diagnostic metrics, suggesting that longer contexts lengths can help, must most discriminative information is captured within the first minute of telemetry.
For full-lead classification, \model{} achieves the best performance on 7 of 17 tasks, including the out-of-domain CSN and \penn{} datasets. Notably, \model{} performs best on 3 of 4 PTB-XL tasks and attains competitive F1 scores on Code15 and HEEDB.

A consistent trend across datasets is that LLM-based models (PULSE, GEM, CAMEL) struggle on tasks with large label spaces, such as HEEDB and CSN (Table \ref{tab:classification-full}), whereas non-LLM architectures (e.g., MELP and MERL) maintain strong performance in these multi-class settings.

\paragraph{Linear probing.}
To evaluate the quality of the latent representation learned by \model{}, we conduct linear probing classification experiments where we freeze the parameters of the model and train a linear layer with 1\% of the dataset's training data. We report the results in Table \ref{tab:probing-full}.
We compare \model{}-Base to the standard baselines as well as the \model{} architecture trained solely on ECGInstruct. 
\model{} outperforms baselines on both in-domain and out-domain datasets across all tasks, demonstrating that it has learned discriminative representation of the ECG signals.
\model{}'s superior performance to the ECGInstruct-only finetuned version demonstrates the usefulness of the curriculum training for learning representations.

\paragraph{Report generation.}
We present the full report generation results in Table \ref{tab:report-full}, and provide an example report generated by \model{} in Figures \ref{fig:report-example} and \ref{fig:report-example-mimic}. Using LLM-as-a-judge, \model{} achieves superior performance on MIMIC-IV. \model{} also comes within 1 point of GEM, the top-performing model on PTB-XL on a 30-point scale.
\model{} also attains competitive scores on the 5 NLP metrics for both MIMIC-IV and PTB-XL, demonstrating strong conversational ability.

\paragraph{Single and Multi-ECG QA.}
We report results for the multiple choice and single-ECG questions from ECG-QA, as well as multi-ECG questions, in Table \ref{tab:comprehensive_eval}.
Because Stages 3 and 4 of the curriculum contain diverse multi-ECG questions, \model{} achieves superior accuracy on multi-ECG questions compared to GEM and PULSE. 
\model{} also maintains strong performance on single-ECG questions in ECG-QA, indicating that training on comparative multi-ECG reasoning does not degrade performance in the single-ECG setting. 

\paragraph{Statistics.}
We report performance on the ECG grounding dataset in Table \ref{tab:comprehensive_eval}. Compared to the baselines, \model{} achieves the lowest average RMSE on statistics computations.
Notably, \model{}'s RMSE is nearly half that of GEM, a model trained on ECG reasoning grounded in statistics.
This improvement likely stems from our staged curriculum: Stage 3 introduces focused, direct statistical computation questions, while Stage 4 integrates these computations into diagnosis tasks. In contrast, GEM encounters statistics only the context of diagnostic reasoning, without an intermediate phase dedicated to learning the underlying computations.

\renewcommand{\arraystretch}{1.2}
\begin{table*}[t]
\centering
\caption{Comprehensive evaluation results across three tasks: Forecasting, Classification, and Report Generation. Best results are highlighted in ``-" indicates method not evaluated on task/dataset.}
\label{tab:comprehensive_eval}
\footnotesize
\setlength{\tabcolsep}{3pt}
\begin{tabular}{l|ccccccc|cc|c|cc}
\toprule
& \multicolumn{7}{c|}{\textbf{Classification}} 
& \multicolumn{2}{c|}{\textbf{ECG-QA}} 
& \multicolumn{1}{c|}{\textbf{Stat}} 
& \multicolumn{2}{c}{\textbf{Report}} \\
\cline{2-13}
& \multicolumn{1}{c}{\textbf{PTB-XL}} 
& \multicolumn{1}{c}{\textbf{Code15}} 
& \multicolumn{1}{c}{\textbf{CPSC}} 
& \multicolumn{1}{c}{\textbf{CSN}} 
& \multicolumn{1}{c}{\textbf{HEEDB}} 
& \multicolumn{1}{c}{\textbf{Icentia}} 
& \multicolumn{1}{c|}{\textbf{\penn{}}} 
& \multicolumn{1}{c}{\textbf{Single}} 
& \multicolumn{1}{c|}{\textbf{Multi}} 
& \multicolumn{1}{c|}{\textbf{CPSC}} 
& \multicolumn{1}{c}{\textbf{PTB-XL}}
& \multicolumn{1}{c}{\textbf{MIMIC}}\\
\cline{2-13}
\multicolumn{1}{l|}{\textbf{Method}} 
& \multicolumn{7}{c|}{\textbf{F1}} 
& \multicolumn{2}{c|}{\textbf{Accuracy}} 
& \multicolumn{1}{c|}{\textbf{RMSE}} 
& \multicolumn{2}{c}{\textbf{LLM}}  \\
\midrule
\textbf{MERL} & 19.22 & 19.72 & 24.86 & 6.38 & 16.33 & 15.80 & 2.53 & - & - & -  & - & - \\ 
\textbf{MELP} & 18.62 & 26.07 & 20.75 & 5.00 & 19.03 & 10.80 & 2.53 & - & - & - & - & -\\ 
\textbf{PULSE} & 28.06 & 75.90 & 57.08 & {12.62} & 18.62 & 25.15 & 34.12 & 79.07  & 64.41 & 144148 & 18.40 & 22.05 \\ 
\textbf{GEM} &29.82 & \textbf{80.83} & \textbf{61.06} & 8.19 & 19.40 & 18.70 & 33.33 & \textbf{80.80}  & 65.89 & 304 & \textbf{20.45} & 44.65 \\
\midrule
\rowcolor[gray]{0.9}\multicolumn{1}{l|}{\textbf{\model{}}} & \textbf{40.37} & 78.20 & 56.12 & \textbf{12.84} & \textbf{20.89} & \textbf{41.14} & \textbf{67.53} & 72.99 & \textbf{69.23} & \textbf{109} & 19.45 & \textbf{62.59} \\
\bottomrule
\end{tabular}
\vspace{0.5em}
\end{table*}

\subsection{Ablation}

We study the design components present in the architecture of \model{}. We vary LoRA adapter usage, special ECG tokens, attentions masks, and train with different combinations on ECGInstruct and evaluate on its counterpart test set ECGBench.  

As shown in Table \ref{tab:ablation-architecture}, the removal of LoRA significantly negatively impacts the performance.
We posit that, due to \model{}'s relatively simple encoding network, most of the trainable parameters stem from LoRA. 
LoRA allows pretrained LLM's transformer blocks to learn deeper and more complex relationships betwen text and ECG.

Finally, we evaluate \model{}'s lead-aware attention masking scheme against full bi-directional attention within each ECG block, and vanilla causal attention throughout all tokens.
As shown in Table \ref{tab:ablation-architecture}, our ECG attention masking scheme outperforms all other options. 
We attribute the benefits of this scheme primarily due to the nature of ECGs: at time $t$, all leads represent the same state of the heart.

\begin{table*}[t]
\centering
\scriptsize
\caption{{Linear probing performance (AUC [\%]) of ECG classification across multiple datasets, 12-leads.} 
Results are reported for different training data proportions (1\%, 10\%, 100\%) \textbf{CAMEL (ECGInstruct)} refers to CAMEL trained only on ECGInstruct data \cite{pulse}, whereas \textbf{CAMEL} refers to the model trained on our curriculum specified in $\S$ \ref{sec:training}.
}
\label{tab:probing-full}
\setlength{\tabcolsep}{2pt}
\begin{tabular}{c|ccc|ccc|ccc|ccc|ccc|ccc}
\toprule
{\textbf{Methods}} & 
 \multicolumn{3}{c|}{\textbf{PTBXL-Rhythm}} & 
 \multicolumn{3}{c|}{\textbf{PTBXL-Sub}} & 
 \multicolumn{3}{c|}{\textbf{PTBXL-Form}} & 
 \multicolumn{3}{c|}{\textbf{PTBXL-Super}} &
 \multicolumn{3}{c|}{\textbf{CPSC2018}} &
 \multicolumn{3}{c}{\textbf{CSN}} \\
 \midrule
{Training Ratio} & 1\% & 10\% & 100\% & 1\% & 10\% & 100\% & 1\% & 10\% & 100\% & 1\% & 10\% & 100\% & 1\%  & 10\% & 100\% & 1\% & 10\% & 100\% \\
\midrule
MERL \citeyearpar{liu2024merl}  & 53.33 & 82.88 & 88.34 & 64.90 & 80.56 & 84.72 & 58.70 & 72.43 & 79.65 & {82.39} & {86.27} & {88.67} & 70.33 & 85.22 & 90.57 & 66.60 & {82.74} & 87.95\\
MELP \citeyearpar{wang2025melp} & {88.83} &  {94.65} & {96.91} & {79.22} & {84.40} & {87.46} & {63.44} & {76.71} & {83.30} & {85.82} & {87.61} & {87.87} & \textbf{88.54} & {91.75} & {94.32} & {78.25} & {84.83} & {90.17}  \\
PULSE \citeyearpar{pulse} & \textbf{98.38} & {99.07} & 99.25 & 89.21 & 92.44 & 93.88 & 82.38 & {90.19} & 91.97 & {89.88} & {92.32} & 93.08 & 67.81 & 73.90 & 82.48 & 87.25 & 89.49 & 91.87 \\
GEM \citeyearpar{gem-empowering-mllm} & 97.55 & 98.78 & 99.12 & 89.48 & {92.57} & 93.19 & {83.32} & 88.07 & 90.77 & 87.61 & 91.88 & 93.03 & 74.58 & 82.41 & 91.38 & 87.12 & 90.33 & 91.69 \\
\rowcolor[gray]{0.9}\textbf{\model{} (ECGInstruct)} & {97.74} & {98.98} & {99.07}   & {90.07}  &	{92.38} & {93.75} & {81.28} &	{89.47} &  {92.17}  &{87.59} & {91.58} & {92.33}	& 	78.54 & 	{93.37}	& {98.52} & {90.00} & {95.05}  & {97.11} \\
\rowcolor[gray]{0.9}\textbf{\model{}} & {98.13} & \textbf{99.17} & \textbf{99.39} & \textbf{90.52} & \textbf{93.41} & \textbf{94.14} & \textbf{85.98} & \textbf{91.14} & \textbf{93.27} & \textbf{91.40} & \textbf{93.73} & \textbf{94.73} & {86.04} & \textbf{96.15} & \textbf{99.40} & \textbf{93.52} & \textbf{97.17} & \textbf{98.36}  \\
\bottomrule
\end{tabular}
\end{table*}

\begin{table*}[t]
    \centering
    \caption{Architecture ablations (F1 [\%]).  Training without LoRA adapters leads to poor ECG representations, as evidenced by the zero F-1 scores on PTB-XL tasks. Lead-aware masking, which enables bidirectionality in attention computation for ECG leads, proves be better, or at-par with full masking.  Both LoRA adapters and lead-aware masking are necessary for achieving SOTA accuracy.  }
    \label{tab:ablation-architecture}
    \footnotesize
    \centering
    \begin{tabular}{l|ccccc|cc}
    \toprule
    \textbf{Method}
        & \makecell[c]{\scriptsize \textbf{PTBXL-Super}}
        & \makecell[c]{\scriptsize\textbf{PTBXL-Sub}}
        & \makecell[c]{\scriptsize\textbf{PTBXL-Rhythm}}
        & \makecell[c]{\scriptsize\textbf{PTBXL-Form}}
        & \makecell[c]{\scriptsize\textbf{Code15}}
        & \makecell[c]{\scriptsize\textbf{CPSC-2018}}
        & \makecell[c]{\scriptsize\textbf{CSN}}
         \\
        \midrule
        \textbf{No LoRA + Lead-aware masking} & 0.00 & 0.00 & 0.00 & 0.00 & 27.35 & 1.18 & 0.00 \\
        \textbf{LoRA + Full masking} & 54.39 & 17.22 & 21.34 & 16.98 & 75.13 & 46.69 & 10.15 \\
        \textbf{LoRA + Causal masking} & 69.15 & 22.13 & 21.08 & 14.72 & 77.34 & 42.30 & 10.02 \\
        \rowcolor[gray]{0.9} \textbf{LoRA + Lead-aware masking} & 75.91 & 26.49 & 34.27 & 19.03 & 81.54 & 51.10 & 14.30 \\
\bottomrule
\end{tabular}
\end{table*}

\section{Related Work}
\label{sec:related}

In this section, we discuss the field of ML-based ECG comprehension and divide it into (i) unimodal ECG foundation models, (ii) multimodal ECG models, including ELMs, (iii) general natural-language-based ECG models, and
(iv) long-horizon cardiac event forecasting.

\paragraph{Unimodal ECG foundation models.}
Recent advances in foundation models have transformed ECG analysis through large-scale pretraining. HuBERT-ECG \citep{coppola2024hubert} uses masked signal reconstruction, predicting masked ECG segments from unmasked context. ECG-FM \citep{mckeen2025ecgfm} combines signal masking with contrastive learning across ECG segments.
ECGFounder \citep{li2025ecgfounder} employs supervised multi-label classification on expert diagnostic annotations. These foundation models are unimodal, i.e.~trained solely on ECG signals during training, and require fine-tuning for downstream tasks.

\paragraph{Multimodal ECG models.}
Multimodal models such as MERL \citep{liu2024merl} enable zero-shot classification through joint learning on ECG records and clinical reports.
MELP \citep{wang2025melp} also trains on both modalities, but adds  hierarchical supervision at the token, beat, and rhythm-level to align ECG with reports at different time-scales.
PULSE \citep{pulse} is another ECG-text model that performs LLaVA-style finetuning for ECG image analysis.
GEM \citep{gem-empowering-mllm} jointly trains ECG and text with a conversational dataset grounded in ECG statistics.
As noted earlier, the inputs to these models are limited to at most 10 seconds.

\paragraph{Natural language generation \& understanding.}
Natural language generation-based methods are more powerful, since they implicitly include classification, and can also support report generation and general QA. Recent approaches are trained in a multi-stage fashion, either via self-supervised or contrastive learning. For example, ECG-ReGen \citep{tang2024ecgregen} uses a self-supervised encoder, ECG-Chat \citep{zhao2025ecgchat} employs contrastive learning for feature alignment, MEIT \citep{wan2025meit} focuses on multimodal instruction tuning, and anyECG-chat \citep{li2025anyecgchat} utilizes three-step curriculum training for variable length inputs. 
Additionally, METS \citep{frozen-LM-zero-shot-learning} uses self-supervised learning to train an ECG encoder paired with a frozen language model.
A notable exception is ECG-Byte \citep{han2025ecgbyte}, which bypasses a specialized encoder by directly tokenizing the ECG signal, enabling end-to-end LLM training.

\paragraph{Forecasting cardiac events.}
Deep learning and classical ML techniques have also demonstrated potential for long-horizon forecasting of adverse cardiac events.
\citet{jhu-picu} apply XGBoost to forecast pediatric cardiac arrest from clinical data, reporting high auROC and auPRC.
\citet{in-hospital-ed-ca-prediction} also use XGBoost in addition to other classical ML models to forecast in-hospital cardiac arrest for emergency department (ED) patients, reporting high auROC.
However, both of these works use clinical features beyond raw ECG, e.g., ED occupancy, limiting their broader applicability.
Conversely, \citet{forecasting-imminent-afib} use convolutional-transformer models for forecasting imminent atrial fibrillation events using only long-term ECG waveforms, demonstrating that such models can learn predictive features from ECG without additional clinical features.

\section{Limitations and Future Work}
\label{sec:limitations}

There are two main limitations with \model's tokenization strategy.
While \model\ can handle ECG segments longer than 10 seconds, the maximum input ECG segment is limited by the context length of the backbone LLM.
Additionally, segmenting ECG into 1-second chunks may truncate the QRS complex or fail to capture subtle morphological changes in the signal.
For future work, we plan to explore different tokenization strategies such as segmenting based on QRS intervals, which could serve as semantic unit of explanation, or using 5 second ECG segments, which would allow longer duration ECG to fit into the context window.

\section{Conclusion}
\label{sec:conclusion}

In this work, we introduce \model, an ECG language model for forecasting cardiac events.
Unlike prior works, \model{} supports flexible ECG context windows exceeding 10 seconds, enabling reasoning over longer signals.
Its multimodal architecture combines a CNN-based ECG encoder and an LLM with LoRA adapters, and is trained using a 5-stage curriculum that progressively builds ECG understanding and forecasting capability.
\model{} achieves SOTA results on prior benchmarks as well as our newly introduced forecasting benchmark, ECGForecastBench.

\begin{ack}
This research was supported by ARPA-H program on Safe and Explainable AI under Award D24AC00253-00, NIH Award R01-EB036016, the Google TPU Builders Program, and a gift from AWS AI to ASSET (Penn Engineering Center on Trustworthy AI). We also thank the Penn Advanced Research Computing Center (PARCC) for providing high-performance computing infrastructure for training CAMEL-ECG.
\end{ack}

\bibliographystyle{iclr2026_conference}
\bibliography{references}


\newpage
\appendix
\appendix

\section{Experimental Set-up}\label{appendix-experimental-set-up}

\begin{table*}[h]
    \centering
    \footnotesize
    \caption{Overview of the ECG Datasets. }
    \label{tab:ecg-overview}
    \begin{tabular}{l|l|l|l|l|l}
    \toprule
    \textbf{Supervision} & \textbf{Dataset Name} & \textbf{Source/Collection} & \textbf{Records} & \textbf{Duration} & \textbf{Leads} \\
    \midrule

    \multirow{5}{*}{\textbf{Signal-only}} 
    & MITDB \citeyearpar{mitdb} 
        & Beth Israel & 48 & 30 m & 2 \\
    & NTUH \citeyearpar{arrest-ntuh-ecg} 
        & National Taiwan University Hospital & 1,000 & 10 s & 12 \\
    & PTB \citeyearpar{ptb}  
        & University Clinic Benjamin Franklin & 549 & 30--120 s & 12 \\
    & QTDB \citeyearpar{qtdb}  
        & Beth Israel, Multi-center Europe & 105 & 15 m & 2 \\
    & SPH \citeyearpar{sph} 
        & Shandong Provincial Hospital & 25,770 & 10--60 s & 12 \\
    & G12 \citeyearpar{challenge2020} 
        & Emory University & 10,344 & 10 s & 12 \\

    \midrule
    \multirow{5}{*}{\makecell[l]{\textbf{Labels}\\\textbf{(in-distribution)}}} 
    & Code-15\% \citeyearpar{code15} 
        & Telehealth Network (Brazil) & 345,779 & 7/10 s & 12 \\
    & HEEDB \citeyearpar{harvard-emory-ecg-database} 
        & Harvard University, Emory University & 11,440,211 & 10 s & 12 \\
    & Icentia11k \citeyearpar{icentia11k} 
        & CardioSTAT (Canada) & 11,000 & 1--2 weeks & 1 \\
    & MIMIC-IV \citeyearpar{mimic-iv-ecg} 
        & Beth Israel & 795,546 & 10 s & 12 \\
    & PTB-XL \citeyearpar{ptbxl} 
        & Schiller AG (Germany) & 21,837 & 10 s & 12 \\
    
    \midrule
    \multirow{5}{*}{\makecell[l]{\textbf{Labels}\\\textbf{(out-of-distribution)}}} 
    & CPSC-2018 \citeyearpar{cpsc2018} 
        & Multi-center China & 13,256 & 6--144 s & 12 \\
    & CSN \citeyearpar{champman} 
        & \makecell[l]{Chapman University, Shaoxing People’s \\ Hospital, Ningbo First Hospital} 
        & 80,938 & 10 s & 12 \\
    & \penn &  \penn{} hospital system & 191 & 10 m - 48 hrs & 8 \\
    \bottomrule
    \end{tabular}
\end{table*}

We collect 1 private and 13 public ECG datasets (listed in Table \ref{tab:ecg-overview}). The signal-only datasets are used only in training the autoencoder, i.e. Stage 1. The in-distribution datasets are used in Stages 2, 3 and 4; we only use the Harvard-Emory dataset in Stages 2 and 3 due to data use policies prohibiting LLM API calls. For each in-distribution dataset, we set aside 20\% of the records for validation and evaluation
and use the remaining records for training. For the out-of-distribution datasets, we do not use \textit{any} samples for training.

\subsection{Hyperparameters} 
\label{appendix:hyperparam}
We list all global and stage-specific hyperparameters in Tables \ref{tab:hyperparams} and \ref{tab:global_hyperparams}.

\begin{table}[h]
    \centering
    \caption{Hyperparameters for Training Stages 2-5}
    \vspace{2mm}
    \begin{tabular}{lcccc}
        \toprule
        \textbf{Stage} & \textbf{Learning Rate} & \textbf{Epochs} & \textbf{Batch Size} & Warmup steps \\
        \midrule
        Stage 2 & 1e-4 & 3 & 8 & 500 \\
        Stage 3 & 1e-4 & 1 & 8 & 500 \\
        Stage 4 & 1e-4 & 9 & 4 & 256 \\
        Stage 5 & 1e-4 & 3 & 2 & 256 \\
        \bottomrule
    \end{tabular}
    \label{tab:hyperparams}
\end{table}
\begin{table}[h]
    \centering
    \caption{Global Hyperparameters}
    \vspace{2mm}
    \begin{tabular}{lc}
        \toprule
        \textbf{Hyperparameter} & \textbf{Value} \\
        \midrule
        Total Gradient Accumulation Target & 4 \\
        Gradient Clipping & 1.0 \\
        Optimizer & AdamW \\
        Learning rate scheduler & linear \\
        LoRA Rank & 128 \\
        Mask strategy & semantic \\
        \bottomrule
    \end{tabular}
    \label{tab:global_hyperparams}
\end{table}

\subsection{Baselines and Tasks}
\label{appendix:classification-setup}

For MERL, we evaluate using \texttt{ncbi/MedCPT-Query-Encoder} text encoder and 1D ResNet18 ECG encoder with embedding dimension 256.
For MELP, we evaluate using \texttt{fuyingw/heart\_bert} text encoder and 8-layer ECG encoder with 12 attention heads, embedding dimension 768, and feed-forward network with embedding dimension 3072.
For PULSE, we use \texttt{PULSE-ECG/PULSE-7B} model. 
For \model, we set \texttt{max\_new\_tokens} = 1000, \texttt{temperature} = 1.0, and \texttt{top\_p} = 0.95. 

We divide PTB-XL classification task into superclass, subclass, rhythm, and form, following prior work \citep{ptbxl}. 
Similarly, for HEEDB classification, we divide into overall interpretation, conduction, ectopy, ischmeia, rhythm, hypertrophy, and infarct sub-tasks.
The ECG-QA dataset contains 3 types of single-ECG questions and 4 types of multi-ECG questions which are sourced from PTB-XL and MIMIC-IV-ECG respectively.

\section{Additional Experimental Results}

\subsection{Forecasting}

\begin{table*}[h]
    \centering
    \caption{Forecasting results (F1 [\%]) on Icentia. The task is forecasting atrial fibrillation or flutter (abnormal) or normal sinus rhythm at varying horizons $h$ (s) and input-windows $w$ (s). We report supervised training results for XGB and CNN, zero-shot results for GPT-5.2, PULSE, GEM, and \model{}, and linear probing result for CAMEL Probe.}
    \label{tab:appendix-forecasting}
    \footnotesize
    \begin{tabular}{l | cccccc | cccccc}
        \toprule
          & \multicolumn{6}{c|}{$h = 60$} &
          \multicolumn{6}{c}{$h = 180$} \\
        \midrule
        \textbf{Method} & 10 & 30 & 60 & 120 & 300 & 600 & 10 & 30 & 60 & 120 & 300 & 600 \\
        \midrule
        \textbf{XGB}  & 59.55 & 55.58 & 56.89 & 56.09 & 59.64 & 55.20 & 56.85 & 56.60 & 58.18 & 57.09 & 58.07 & 56.17 \\
        \textbf{CNN} & 46.89 & 54.62 & 53.59 & 50.79 & 52.35 & 53.10 & 49.59 & 54.72 & 57.08 & 54.40 & 52.76 & 50.21 \\
        \rowcolor[gray]{0.9}\textbf{\model{} Probe} & \textbf{69.83} & \textbf{72.52} & \textbf{69.43} & \textbf{74.30} & 72.56 & \textbf{75.14} & \textbf{69.81} & \textbf{67.38} & \textbf{69.91} & \textbf{69.85} & \textbf{75.46} & \textbf{74.64} \\
        \midrule
        \textbf{GPT-5.2} & 43.26 & 47.54 & 48.00 & 48.24 & 54.36 & 51.28 & 37.67 & 46.71 & 51.22 & 54.43 & 49.29 & 52.01 \\
        {\textbf{PULSE}} & 43.21 & - & - & - & - & - & 40.50 & - & - & - & - & -  \\
        {\textbf{GEM}} & 32.54 & - & - & - & - & - & 32.95 & - & - & - & - & -\\
        \rowcolor[gray]{0.9}\textbf{\model{}} & 63.28 & 68.88 & 69.23 & 71.37 & \textbf{73.42} & 73.18 & 60.04 & 61.27 & 65.44 & 69.59 & 71.34 & 67.53 \\
        \bottomrule
    \end{tabular}
    
    \vspace{3mm}
    \begin{tabular}{l | cccccc | cccccc}
        \toprule
          & \multicolumn{6}{c|}{$h = 300$} &
          \multicolumn{6}{c}{$h = 600$} \\
        \midrule
        \textbf{Method} & 10 & 30 & 60 & 120 & 300 & 600 & 10 & 30 & 60 & 120 & 300 & 600 \\
        \midrule
        \textbf{XGB}& 54.08 & 54.97 & 54.35 & 57.05 & 56.86 & 53.37 & 57.93 & 57.12 & 53.84 & 55.06 & 56.03 & 57.07\\
        \textbf{CNN} & 48.47 & 53.13  & 46.54 & 46.32 & 57.72 & 48.99 & 48.85 & 51.96 & 53.69 & 51.32 & 49.90 & 56.19\\
        \rowcolor[gray]{0.9}\textbf{\model{} Probe} & \textbf{68.54} & \textbf{71.05} & {69.31} & \textbf{70.30} & \textbf{73.36} & \textbf{72.70} & \textbf{69.30} & \textbf{69.53} & \textbf{68.49} & \textbf{72.04} & \textbf{73.54} & \textbf{76.02} \\
        \midrule
        \textbf{GPT-5.2} & 43.14 & 53.42 & 52.07 & 51.19 & 58.16 & 56.37 & 46.11 & 56.04 & 47.55 & 55.80 & 50.01 & 55.41 \\
        {\textbf{PULSE}}  & 36.67 & - & - & - & - & - & 40.12 & - & - & - & - & - \\
        {\textbf{GEM}} & 33.33 & - & - & - & - & - & 31.49 & - & - & - & - & -\\
        \rowcolor[gray]{0.9}\textbf{\model{}} & 58.07 & 63.58 & \textbf{70.15} & 67.15 & 70.73 & 70.48 & 58.54 & 63.91 & 64.28 & 65.99 & 71.37 & 66.90 \\
        \bottomrule
    \end{tabular}
\end{table*}

\subsection{Input Classification}

We report per-task classification results for 10-second 12-lead ECG signals in Table \ref{tab:classification-full}. 
Additionally, we compare the performance of classifying AF across single lead rhythm strips ranging in duration from 10 seconds to 15 minutes (Table \ref{tab:appendix-length}) and using only a fixed selection of 6, 4, 3, 2, and 1 leads (Table \ref{tab:classification-fixed}). 

\clearpage
\begin{table*}[h]
    \centering
    \caption{Full classification results (F1 [\%]).}
    \label{tab:classification-full}
    \footnotesize
    \centering
    \begin{tabular}{l|cccc|cc}
    \toprule
    \textbf{Method}
        & \makecell[c]{\textbf{PTBXL}\\\textbf{Super}}
        & \makecell[c]{\textbf{PTBXL}\\\textbf{Sub}}
        & \makecell[c]{\textbf{PTBXL}\\\textbf{Rhythm}}
        & \makecell[c]{\textbf{PTBXL}\\\textbf{Form}}
        & \makecell[c]{\textbf{Code15}\\\textbf{Binary}}
        & \makecell[c]{\textbf{Code15}\\\textbf{Multi}}\\
    \midrule
    \textbf{MERL} 
        & 41.39 & 12.36 & 10.84 & {12.31} & 3.09 & 36.35   \\
    \textbf{MELP} 
        & 40.45 & 10.45 & 11.43 & 12.16 & 20.50 & 31.63  \\
    \textbf{PULSE} & 73.47 & 8.11 & 18.07 & 12.59 & 62.28 & 89.52   \\
    \textbf{GEM} & \textbf{75.77} & 8.01 & 20.87 & 14.64  & \textbf{71.12} & \textbf{90.53} \\
    \rowcolor[gray]{0.9}\textbf{\model } & 67.75 & \textbf{28.89} & \textbf{39.29} & \textbf{25.54} & 69.45 & 86.95 \\
    \bottomrule
    \end{tabular}
    
 \vspace{3mm}
     \begin{tabular}{l|ccccccc}
     \toprule
     \textbf{Method}
      & \makecell[c]{\textbf{HEEDB}\\\textbf{Binary}}
         & \makecell[c]{\textbf{HEEDB}\\\textbf{Conduction}}
         & \makecell[c]{\textbf{HEEDB}\\\textbf{Ectopy}}
         & \makecell[c]{\textbf{HEEDB}\\\textbf{Ischemia}}
         & \makecell[c]{\textbf{HEEDB}\\\textbf{Rhythm}}
         & \makecell[c]{\textbf{HEEDB}\\\textbf{Hypertrophy}}
         & \makecell[c]{\textbf{HEEDB}\\\textbf{Infarct}}
     \\
     \midrule
     \textbf{MERL}  & 4.27 & \textbf{18.05} & 17.41 & \textbf{17.21} & \textbf{10.86} & \textbf{20.67} & \textbf{25.81} \\
     \textbf{MELP}  & 24.96 & 16.56 & \textbf{18.26} & 16.97 & 10.62 & 20.42 & 25.43 \\
     \textbf{PULSE}  & 86.36 & 14.15 & 3.33 & 4.59 & 5.74 & 7.73 & 8.44 \\
     \textbf{GEM} & 88.95 & 14.15 & 3.48 & 6.41 & 5.97 & 6.79 & 10.05\\
     \rowcolor[gray]{0.9}\textbf{\model}  & \textbf{96.41} & 13.78 & 3.61 & 6.69 & 10.05 & 11.45 & 4.21 \\
     \bottomrule
     \end{tabular}
\end{table*}

\begin{table}[H]
    \centering
    \caption{AFIB classification results (F1 [\%]) with varying ECG durations (s)}
    \label{tab:appendix-length}
    \vspace{3mm}
    \footnotesize
    \setlength{\tabcolsep}{3pt}
    \begin{subtable}[ht]{\textwidth}
    \centering
    \begin{tabular}{l|ccccccccc}
    \toprule
    \multicolumn{1}{l|}{\textbf{Method}} & 10 & 30 & 60 & 120 & 300 & 600 & 900\\
    \midrule
    \textbf{CAMEL} & 46.19 & 41.32 & 40.45 & 45.10 &  58.28 & 64.95 & 64.15 \\
    \bottomrule
    \end{tabular}
    \end{subtable}
\end{table}

\begin{table}[H]
\vspace{-1mm}
\centering
\caption{Classification results (F1 [\%]) with fewer leads at test time.}
\label{tab:classification-fixed}
\footnotesize
\setlength{\tabcolsep}{3pt}

\begin{subtable}[ht]{\textwidth}
\centering
\caption{Leads I, II, III, aVR, aVL, and aVF.}
    \begin{tabular}{l|cccccccc}
    \toprule
    \textbf{Method}
        & \makecell[c]{\scriptsize \textbf{PTBXL-Super}}
        & \makecell[c]{\scriptsize\textbf{PTBXL-Sub}}
        & \makecell[c]{\scriptsize\textbf{PTBXL-Rhythm}}
        & \makecell[c]{\scriptsize\textbf{PTBXL-Form}}
        & \makecell[c]{\scriptsize\textbf{CPSC-2018}}
        & \makecell[c]{\scriptsize\textbf{CSN}}
        & \makecell[c]{\scriptsize\textbf{Code15-Binary}}
        & \makecell[c]{\scriptsize\textbf{Code15-Multi}} \\
    \midrule
\textbf{MERL} & 43.15 & 12.48 & 11.23 & 12.95 & 22.52 & 4.87 & 45.31 & 33.14 \\
\textbf{MELP} & 41.35 & 12.09 & 10.81 & 12.52 & 20.90 & 5.67 & 41.57 & 33.65 \\
\textbf{PULSE} & 48.48 & 8.29 & 16.47 & 11.48 & 54.71 & 6.65 & 71.34 & 78.97 \\
\textbf{GEM} & 47.75 & 8.96 & 11.69 & 12.03 & 54.86 & 6.67 & \textbf{74.00} & 83.67 \\
\rowcolor[gray]{0.9}{\textbf{\model}} & \textbf{60.79} & \textbf{24.35} & \textbf{41.89} & \textbf{21.42} & \textbf{55.46} & \textbf{10.58} & 69.48 & \textbf{87.99} \\
    \bottomrule
    \end{tabular}
\end{subtable}

\begin{subtable}[ht]{\textwidth}
\centering
\caption{Leads I, II, III, and V2.}
    \begin{tabular}{l|cccccccc}
    \toprule
    \textbf{Method}
        & \makecell[c]{\scriptsize \textbf{PTBXL-Super}}
        & \makecell[c]{\scriptsize\textbf{PTBXL-Sub}}
        & \makecell[c]{\scriptsize\textbf{PTBXL-Rhythm}}
        & \makecell[c]{\scriptsize\textbf{PTBXL-Form}}
        & \makecell[c]{\scriptsize\textbf{CPSC-2018}}
        & \makecell[c]{\scriptsize\textbf{CSN}}
        & \makecell[c]{\scriptsize\textbf{Code15-Binary}}
        & \makecell[c]{\scriptsize\textbf{Code15-Multi}} \\
    \midrule
\textbf{MERL} & 40.60 & 11.96 & 10.96 & 13.02 & 24.16 & 4.15 & 33.54 & 34.56 \\
\textbf{MELP} & 40.14 & 10.87 & 11.00 & 12.91 & 21.44 & 4.46 & 34.58 & 32.48 \\
\textbf{PULSE} & 52.71 & 6.82 & 13.07 & 11.91 & 57.03 & 9.05 & 69.41 & 83.64 \\
\textbf{GEM} & 44.60 & 4.76 & 11.08 & 12.09 & 50.96 & 6.40 & \textbf{72.02} & 78.94 \\
    \rowcolor[gray]{0.9}{\textbf{\model}} & \textbf{67.39} & \textbf{19.15} & \textbf{39.75} & \textbf{23.30} & \textbf{57.78} & \textbf{11.33} & 70.21 & \textbf{89.15} \\
    \bottomrule
    \end{tabular}
\end{subtable}

\begin{subtable}[ht]{\textwidth}
\centering
\caption{Leads I, II, and III.}
    \begin{tabular}{l|cccccccc}
    \toprule
    \textbf{Method}
        & \makecell[c]{\scriptsize \textbf{PTBXL-Super}}
        & \makecell[c]{\scriptsize\textbf{PTBXL-Sub}}
        & \makecell[c]{\scriptsize\textbf{PTBXL-Rhythm}}
        & \makecell[c]{\scriptsize\textbf{PTBXL-Form}}
        & \makecell[c]{\scriptsize\textbf{CPSC-2018}}
        & \makecell[c]{\scriptsize\textbf{CSN}}
        & \makecell[c]{\scriptsize\textbf{Code15-Binary}}
        & \makecell[c]{\scriptsize\textbf{Code15-Multi}} \\
    \midrule
\textbf{MERL} & 40.47 & 12.18 & 10.91 & 12.95 & 24.18 & 3.95 & 33.46 & 34.76 \\
\textbf{MELP} & 40.31 & 11.28 & 11.17 & 12.05 & 22.04 & 5.10 & 37.40 & 34.34 \\
\textbf{PULSE} & 51.97 & 6.35 & 13.26 & 14.39 & 54.89 & 8.71 & 68.24 & 84.14 \\
\textbf{GEM} & 43.18 & 7.17 & 8.74 & 13.92 & 50.26 & 7.09 & \textbf{72.24} & 77.21 \\
    \rowcolor[gray]{0.9}{\textbf{\model}} & \textbf{67.47} & \textbf{18.85} & \textbf{38.03} & \textbf{25.00} & \textbf{59.33} & \textbf{10.28} & 69.38 & \textbf{89.48} \\
    \bottomrule
    \end{tabular}
\end{subtable}
\end{table}

\begin{table}[H]
\footnotesize
\setlength{\tabcolsep}{3pt}

\begin{subtable}[ht]{\textwidth}
\centering
\setcounter{subtable}{3} 
\caption{Leads I and II.}
    \begin{tabular}{l|cccccccc}
    \toprule
    \textbf{Method}
        & \makecell[c]{\scriptsize \textbf{PTBXL-Super}}
        & \makecell[c]{\scriptsize\textbf{PTBXL-Sub}}
        & \makecell[c]{\scriptsize\textbf{PTBXL-Rhythm}}
        & \makecell[c]{\scriptsize\textbf{PTBXL-Form}}
        & \makecell[c]{\scriptsize\textbf{CPSC-2018}}
        & \makecell[c]{\scriptsize\textbf{CSN}}
        & \makecell[c]{\scriptsize\textbf{Code15-Binary}}
        & \makecell[c]{\scriptsize\textbf{Code15-Multi}} \\
    \midrule
\textbf{MERL} & 44.24 & 13.09 & 11.20 & 13.32 & 23.81 & 5.55 & 46.39 & 35.21 \\
\textbf{MELP} & 40.30 & 11.33 & 10.76 & 13.31 & 20.83 & 5.01 & 31.53 & 33.23 \\
\textbf{PULSE} & 43.91 & 5.32 & 13.90 & 11.23 & 51.82 & 4.25 & 66.50 & 72.63 \\
\textbf{GEM} & 43.71 & 9.06 & 5.65 & 13.25 & 43.99 & 7.52 & \textbf{71.22} & 57.07 \\
    \rowcolor[gray]{0.9}{\textbf{\model}}  & \textbf{62.24} & \textbf{17.80} & \textbf{38.96} & \textbf{20.80} & \textbf{56.76} & \textbf{11.93} & 69.86 & \textbf{88.02} \\
    \bottomrule
    \end{tabular}
\end{subtable}

\begin{subtable}[ht]{\textwidth}
\centering
\caption{Lead II.}
    \begin{tabular}{l|ccccccccc}
    \toprule
    \textbf{Method}
        & \makecell[c]{\scriptsize \textbf{PTBXL-Super}}
        & \makecell[c]{\scriptsize\textbf{PTBXL-Sub}}
        & \makecell[c]{\scriptsize\textbf{PTBXL-Rhythm}}
        & \makecell[c]{\scriptsize\textbf{PTBXL-Form}}
        & \makecell[c]{\scriptsize\textbf{CPSC-2018}}
        & \makecell[c]{\scriptsize\textbf{CSN}}
        & \makecell[c]{\scriptsize\textbf{Code15-Binary}}
        & \makecell[c]{\scriptsize\textbf{Code15-Multi}}\\
    \midrule
\textbf{MERL} & 45.06 & 12.98 & 11.39 & 12.71 & 23.64 & 4.66 & 52.29 & 35.49 \\
\textbf{MELP} & 42.09 & \textbf{13.08} & 10.88 & 13.73 & 21.06 & 4.89 & 46.39 & 33.20 \\
\textbf{PULSE} & 38.77 & 5.34 & 12.05 & 11.68 & 41.65 & 5.83 & 67.59 & 59.52 \\
\textbf{GEM} & 44.26 & 6.59 & 5.97 & 10.33 & 34.90 & \textbf{8.48} & \textbf{70.27} & 50.06 \\
\rowcolor[gray]{0.9}{\textbf{\model}} & \textbf{52.79} & 12.40 & \textbf{35.31} & \textbf{19.09} & \textbf{53.54} & 7.97 & 69.25 & \textbf{82.03} \\
    \bottomrule
    \end{tabular}
\end{subtable}
\end{table}

\subsection{{Question Answering}}
\begin{table}[H]
    \centering
    \caption{Full ECG-QA results (Accuracy [\%]) on PTB-XL dataset.}
    \label{tab:ecgqa-full}
    \vspace{3mm}
    \footnotesize
\begin{subtable}[t]{\textwidth}
    \centering
    \begin{tabular}{l|ccc|cccc}
    \toprule
    \textbf{Method}
    & \makecell[c]{Single \\ Choose}
     & \makecell[c]{Single \\ Query}
     & \makecell[c]{Single \\ Verify}
    & \makecell[c]{Multi \\ CQ}
     & \makecell[c]{Multi \\ CV}
     & \makecell[c]{Multi \\ IQ}
     & \makecell[c]{Multi \\ IV}
    \\
    \midrule
    \textbf{PULSE} & 82.22 & 78.07 & 76.91 & 67.96 & 53.75 & 84.71 & 51.21 \\
    \textbf{GEM} & 83.67 & 78.19 & 80.54 &  67.94 & 57.36 & 84.61 & 53.64 \\
    \textbf{\model} & 70.69 & 72.38 & 75.89 & 67.83 & 56.85 & 84.17 & 68.06 \\
    \bottomrule
    \end{tabular}
\end{subtable}
\end{table}

\subsection{Report Generation}
\begin{figure}[H]
  \centering
  {
  \begin{minipage}{0.9\linewidth}
  \centering
    \begin{llmprompt}
    \centering
    Write a diagnostic report for this ECG waveform.
    \begin{figure}
    \vspace{-1mm}
    \centering
      \includegraphics[width=0.4\linewidth]{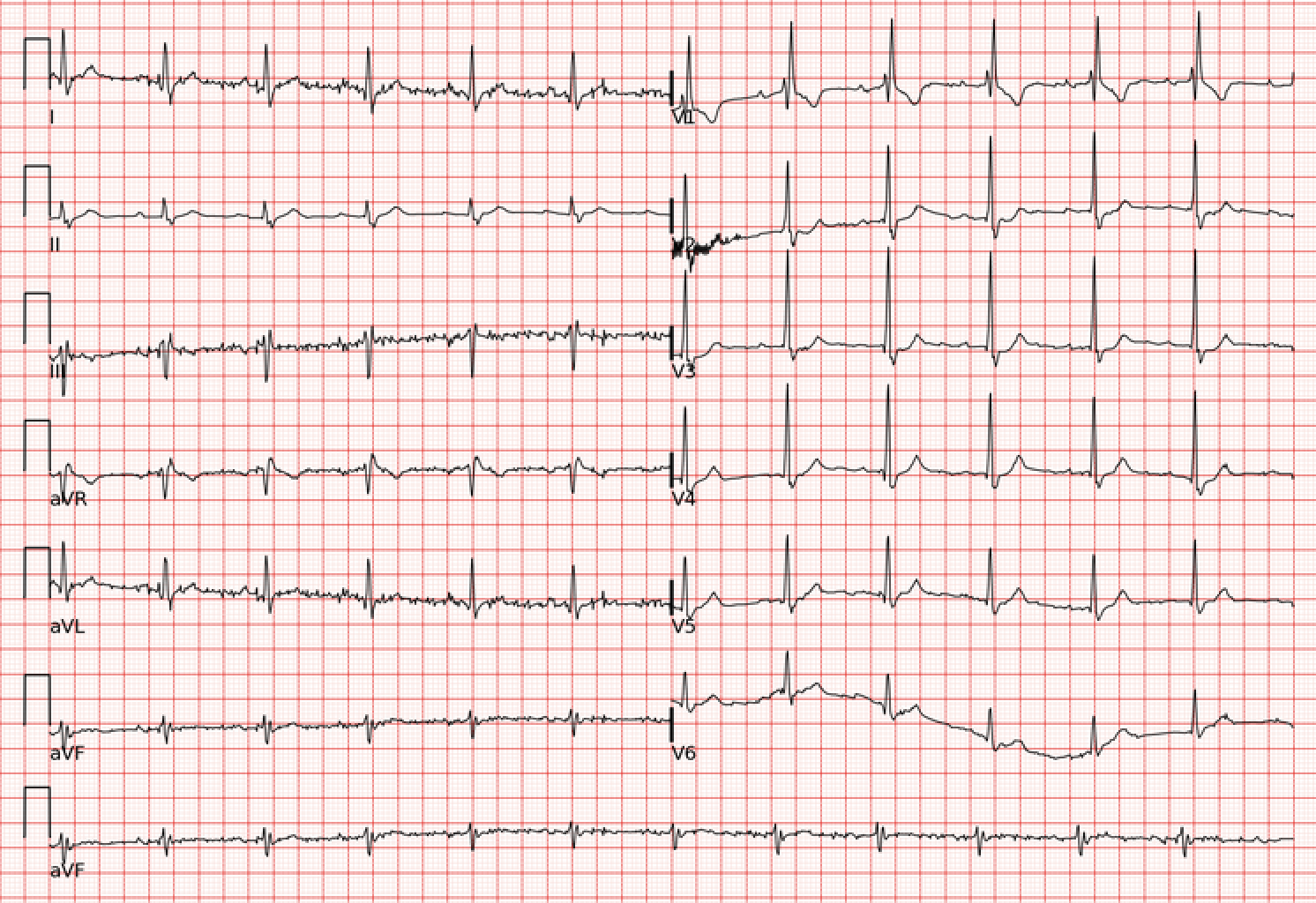}
    \end{figure}
    \end{llmprompt}
    \begin{llmresponse}
    \footnotesize
    The mean RR interval is 819 ms, resulting in a heart rate of 73 bpm (60,000 / 819). The PR interval is 227 ms and the QRS duration is 113 ms. Diagnosis: Sinus rhythm with incomplete right bundle branch block.
\end{llmresponse}
  \end{minipage}
  }
  \caption{Example report generated by \model{} for a sample from the PTB-XL dataset. The ground truth label is \texttt{Sinus rhythm. Incomplete right bundle branch block. PR interval is at the upper limit. Otherwise, normal ECG.}}
  \label{fig:report-example}
\end{figure}

\begin{figure}[H]
  \centering
  {
  \begin{minipage}{0.9\linewidth}
  \centering
    \begin{llmprompt}
    Interpret the provided ECG waveform, identify key features and abnormalities in each lead, and generate a clinical diagnosis that is supported by the observed evidence.
    \begin{figure}
    \vspace{-1mm}
    \centering
    \includegraphics[width=0.5\linewidth]{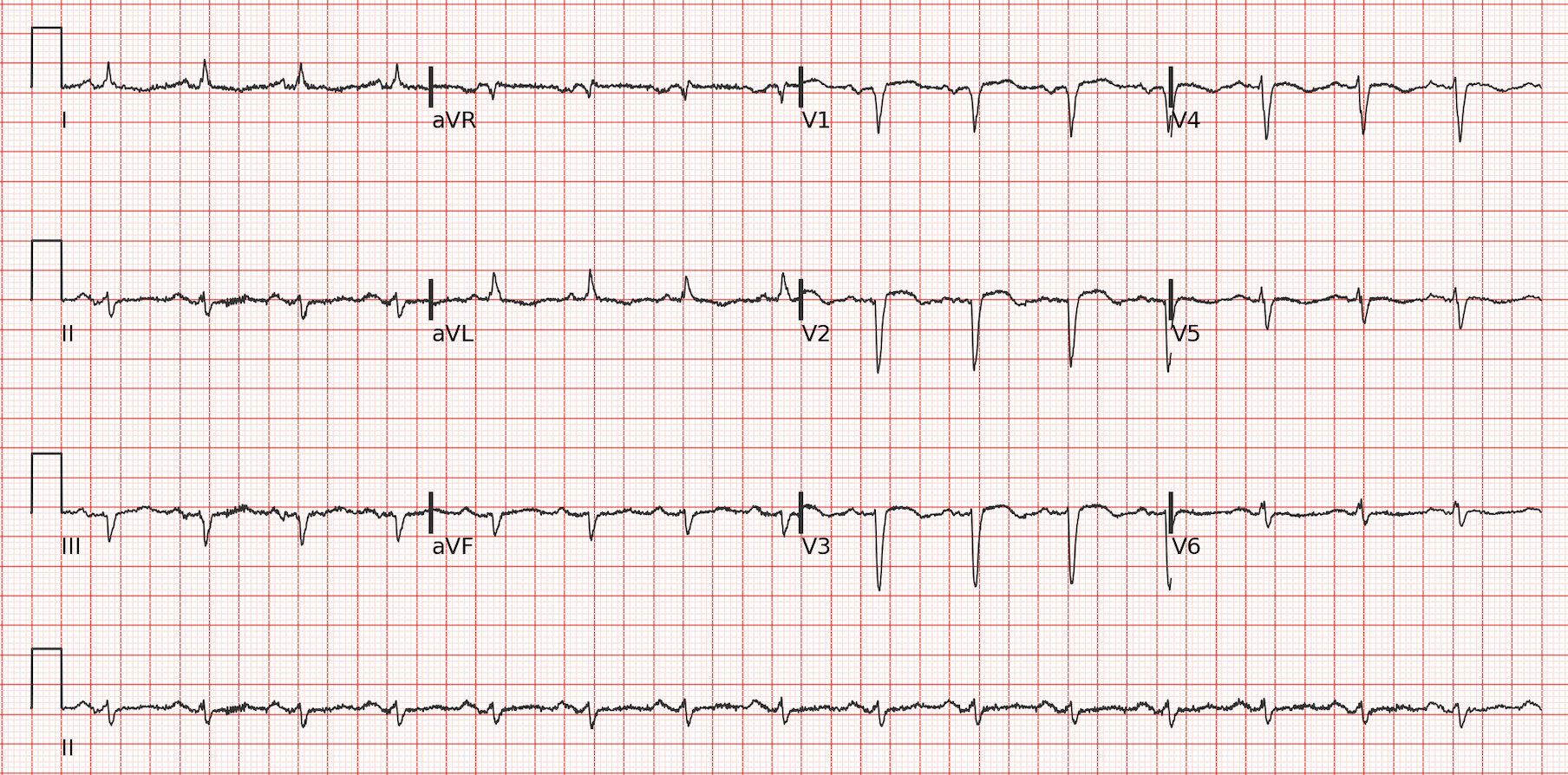}
    \end{figure}
    \end{llmprompt}
    \begin{llmresponse}
    \footnotesize
    To determine the heart rate, we calculate the mean of the RR intervals: (648 + 652 + 652 + 652 + 652 + 652 + 652 + 652 + 648 + 652 + 656 + 652 + 652 + 656 + 652) / 15 = 652 ms. The heart rate is 60,000 / 652 $\approx$ 92 bpm. The rhythm is sinus, with a PR interval of 188 ms and a P-wave duration of 152 ms. The QRS duration is significantly prolonged at 152 ms globally, with Lead V1 specifically showing a duration of 172 ms. The QRS axis is shifted to the left. In the precordial leads, there is a notable loss of R-wave progression and deep negative deflections in V1 (-0.72 mV), V2 (-1.00 mV), and V3 (-1.22 mV), which is indicative of an anteroseptal infarct of undetermined age. Furthermore, the T-wave amplitudes in the lateral leads are negative (V5: -0.07 mV, V6: -0.05 mV), suggesting lateral ischemia.

    $|$ Feature $|$ Value/Interpretation $|$
    
    $|$ Heart Rate $|$ 92 bpm (Sinus) $|$
    
    $|$ PR Interval $|$ 188 ms (Normal) $|$
    
    $|$ QRS Duration $|$ 152 ms (Widened) $|$
    
    $|$ QRS Axis $|$ Left Axis Deviation $|$
    
    $|$ V1-V3 Morphology $|$ Anteroseptal infarct (Undetermined age) $|$
    
    $|$ Lateral T-waves $|$ Inverted in V5-V6 (Ischemia) $|$ 
   
   Diagnosis: Sinus rhythm, Left axis deviation, Anteroseptal infarct - age undetermined, Lateral T wave changes may be due to myocardial ischemia, Abnormal ECG. 
\end{llmresponse}
  \end{minipage}
  }
  \caption{Example of a report generated by \model{} for a sample from the MIMIC-IV dataset. 
  }
  \label{fig:report-example-mimic}
\end{figure}

We present results on evaluating reports using GPT-5.2 as a judge and NLP metrics in Table \ref{tab:report-full}. We show examples of reports generated by \model{} for datasets PTB-XL (Figure \ref{fig:report-example}) and MIMIC-IV (Figure \ref{fig:report-example-mimic}).

\renewcommand{\thefootnote}{\fnsymbol{footnote}} 
\begin{table}[H]
    \caption{Full report generation result.}
    \label{tab:report-full}
    \footnotesize
    \begin{subtable}[t]{0.98\linewidth}
    \caption{PTB-XL}
    \label{tab:report-ptbxl}
    \centering
    \begin{tabular}{l|c|ccccc}
    \toprule
    \multicolumn{1}{l|}{\textbf{Method}} & LLM & BLEU-1 & BLEU-4 & METEOR & ROUGE & BERT F1\\
    \midrule
    \textbf{MELP} & \footnotemark & {13.02} & \textbf{1.87} & 11.28 & \textbf{18.50} & 44.08 \\ 
    \textbf{PULSE} & 18.40 & 9.37 & 1.44 & 24.53 & 9.91 & 83.95 \\
    \textbf{GEM} & \textbf{20.45} & 6.05 & 1.30 & 23.70 & 8.02 & 83.47 \\
    \rowcolor[gray]{0.9}{\textbf{\model}} & 19.45 & \textbf{14.28} & 1.52 & \textbf{31.31} & 10.64 & \textbf{84.00} \\
    \bottomrule
    \end{tabular}
    \end{subtable}
    \vspace{3mm}

\begin{subtable}[t]{0.98\linewidth}
    \centering
    \footnotesize
    \caption{MIMIC-IV.}
        \label{tab:report-mimic-groudning}
    \begin{tabular}{l|c|ccccc}
    \toprule
    \multicolumn{1}{l|}{\textbf{Method}} & LLM & BLEU-1 & BLEU-4 & METEOR & ROUGE & BERT F1\\
    \midrule
    \textbf{PULSE} & 40.43 & 6.09 & 0.93 & 20.74 & 8.02 & 83.37 \\
    \textbf{GEM} & 44.65 & \textbf{50.81} & \textbf{11.60} & \textbf{35.06} & \textbf{23.81} & \textbf{87.78}  \\
    \rowcolor[gray]{0.9}{\textbf{\model}} & \textbf{62.59} &  23.11 & 3.30 & 17.57 & 15.56 & 82.10 \\
    \bottomrule
    \end{tabular}
    \end{subtable}
\end{table}
\footnotetext{Unable to reproduce the result.}
\renewcommand{\thefootnote}{\arabic{footnote}} 

\newpage

\clearpage

\section{Training Dataset}

\subsection{Stage 3} \label{appendix-stage3}

For Stage 3, we generate conversations that require understanding ECG statistics. We use the ECGDeli library \citep{ecgdeli} to compute lead-specific and global statistics for each sample of the datasets from Stage 2.
These statistics include P-QRS-T amplitudes, beat-wise R-R intervals, QRS durations, PR-intervals, ST-segment deviations, ST-slope, and global heart rate metrics.

\begin{figure}[H]
    \centering
    \includegraphics[width=\textwidth]{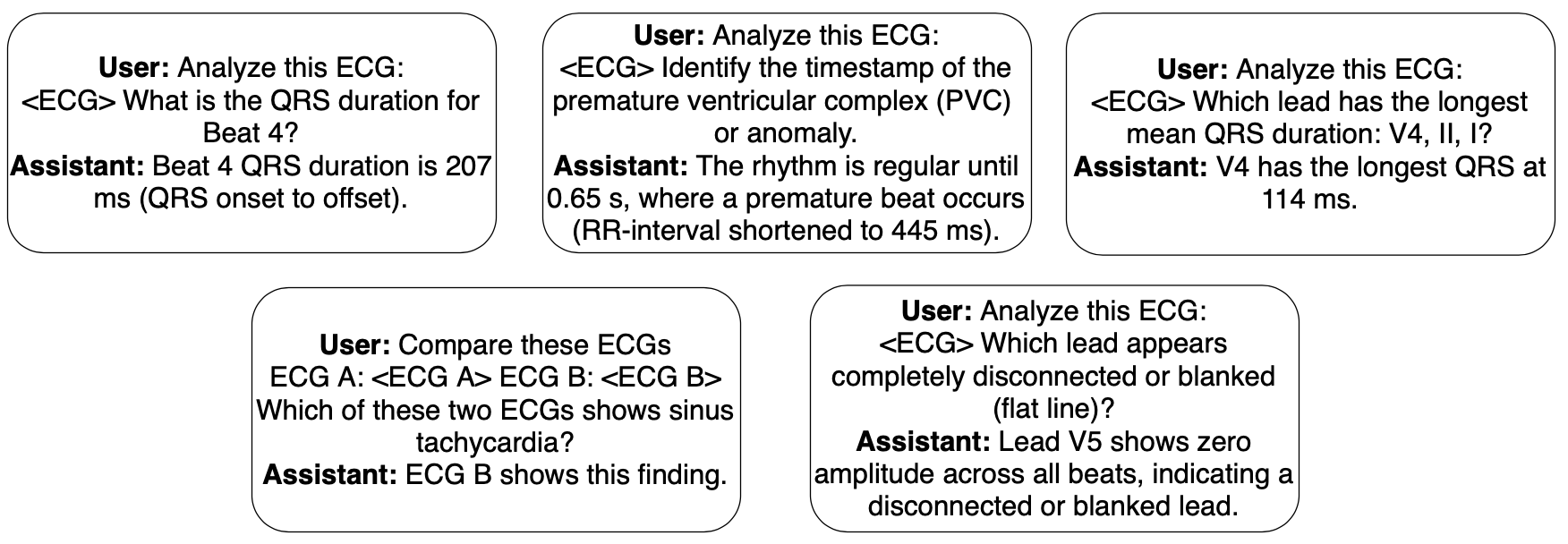}
    \caption{Stage 3 conversation examples across the 5 task types.}
    \label{fig:stage3-questions}
\end{figure}

With these statistics, we use a diverse set of question types to reflect real-world clinical workflows.
Example question types include beat-wise statistics retrieval (e.g., verifying specific wave durations), temporal anomaly detection (e.g, identifying premature beats), and comparative analysis (e.g., comparing feature across two ECGs).
Answer formats include multiple choice, short-answer, and step-by-step explanations.
We provide several examples in Figure \ref{fig:stage3-questions}.
We generate approximately 30 million question-answer pairs from ECG samples spanning the HEEDB, Code-15\%, PTBXL, and MIMIC-IV datasets.

\subsection{Stage 4} \label{appendix-stage4}

The goal of Stage 4 is to generate multi-turn conversational data that has either single-ECG analysis questions or comparison questions.
These conversations use statistics learned in Stage 3 as a part of their reasoning.
We generate approximately 420K samples with ECGs sourced from MIMIC, PTBXL, and Code15.
To generate the first 20K of these conversations, we follow a similar approach as GEM \citep{gem-empowering-mllm}. Specifically, we take the task types and label formats from ECGInstruct \citep{pulse} and use \texttt{gemini-3-flash-preview} to generate reasoning (grounded in ECG statistics) along with the final answer.
To generate the remaining 400K of these conversations, we again use \texttt{gemini-3-flash-preview} with the following system prompt (abridged):

\begin{bigllmprompt}
You are an expert cardiologist. Generate a realistic clinical dialogue interpreting an ECG according to the specifications below. Always include a clinically supported diagnosis consistent with the report and provided statistics.

\paragraph{Objectives}
\begin{enumerate}
    \item Realistic clinical reasoning: Reflect how a cardiologist visually analyzes an ECG and arrives at a diagnosis.
    \item Evidence-based diagnosis: Support conclusions with ECG findings and statistics when applicable. Do not cite statistics that contradict the diagnosis.
    \item Logical flow: Present reasoning from basic observations to final diagnosis.
\end{enumerate}

\paragraph{Dialogue constraints}
\begin{itemize}
    \item Knowledge asymmetry: The human can see the ECG image only and does not know the statistics or diagnosis.
    \item ECG-dependent: The conversation must require the ECG; avoid questions answerable from text alone.
    \item Role discipline: The human asks questions; the \texttt{gpt} role ONLY answers and explains. GPT must NEVER ask questions.
    \item Human limitations: The human must NOT describe ECG features, morphology, or measurements.
    \item Multi-turn continuity: Each turn should build on prior discussion.
\end{itemize}

\paragraph{Clinical reasoning rules}
\begin{itemize}
    \item Reasoning before conclusions: Each \texttt{gpt} turn must include clinical reasoning before any final answer.
    \item Numeric fidelity: All values must exactly match ground truth, including units and lead-specific details when relevant.
    \item Stepwise computation: When interpreting metrics (e.g., heart rate, intervals), show individual measurements and math before the final value.
    \item No patient history assumptions: Discuss only what is observable from the ECG.
    \item No software references: Interpretation must be framed as human visual analysis only.
    \item Avoid filler language; be direct and clinical.
\end{itemize}

\paragraph{Formatting rules}
\begin{itemize}
    \item Return ONLY a JSON list of turns using DOUBLE QUOTES.
    \item Valid elements: \{"from":"human","value":"..."\} or \{"from":"gpt","value":"..."\}.
    \item Do not refer to the assistant as ``doctor'' or ``doc''.
    \item If a specific output format (e.g., SOAP) is required, it must be explicitly requested by the human.
\end{itemize}

\paragraph{Ground truth inputs}
Stats: \{stats\_str\} \\
Diagnosis: \{condition\}

\{blueprint\_spec\}

\textbf{Generate the conversation now:}
    
\end{bigllmprompt}

In order to promote greater diversity in the data, we sample from a set of blueprints, including 28 task types, 10 audiences, 13 formats, and 13 dialogue arcs. We list several examples below:

\begin{blueprints}

\paragraph{Task Types}
\begin{enumerate}
    \item Concise summary: provide a brief clinical summary of the ECG findings
    \item Structured report: generate a structured clinical report with specific fields
    \item QA probing: answer specific questions about ECG features
    \item Clinical explanation: explain the clinical significance of findings
    \item Teaching: teach ECG interpretation concepts using this example
\end{enumerate}

\paragraph{Audiences}
\begin{enumerate}
    \item Clinician-to-clinician: attending physician to fellow/resident
    \item Clinician to patient: doctor explaining to a patient
    \item Teaching trainee: attending teaching a medical student or junior resident
    \item Paramedic handoff: paramedic to ED physician handoff
    \item Researcher note: research annotation or case documentation
\end{enumerate}

\paragraph{Format}
\begin{enumerate}
    \item Narrative: free-form narrative text
    \item Bullet list: bulleted list of findings
    \item Soap note: SOAP note structure
    \item Impression findings: findings then impression format
    \item Short answer rationale: evidence and rationale followed by a brief answer
\end{enumerate}

\paragraph{Dialogue Arcs}
\begin{enumerate}
    \item Direct response: human asks, GPT answers directly
    \item Context volunteered: human volunteers asks whether the diagnosis or explanation would change with additional clinical context
    \item Challenge response: human challenges GPT's interpretation
    \item Lay translation: human asks for simpler explanation
    \item Next steps: human asks about clinical next steps
\end{enumerate}

\paragraph{Audiences}

\end{blueprints}

We list several examples of conversations produced in Stage 4 below.

\begin{multi-turn-conversation}
\textbf{User:} How does this ECG's rhythm compare to a typical normal baseline rhythm, and what is your diagnosis? \\

\textbf{Assistant:} Reasoning: RR intervals (727, 531, 879, 926, 828, 484, 930, 703, 527, 859, 504, 562, 875 ms) show extreme variability (std 168 ms) and absent P waves compared to a steady normal baseline.

Comparison: This rhythm is irregular with a heart rate of 84 bpm, unlike a normal P-wave-driven sinus rhythm.
Diagnosis: Atrial Fibrillation. \\

\textbf{User:} Based on these findings, categorize the clinical priorities for the next shift into emergent, urgent, and routine categories. \\

\textbf{Assistant:} Reasoning: The rhythm is stable Atrial Fibrillation with a controlled rate (84 bpm) and a QRS duration of 115 ms.
\begin{enumerate}
    \item Emergent: None; monitor for rapid ventricular response or hypotension.
    \item Urgent: Cardiology consult for anticoagulation and rate/rhythm strategy.
    \item Routine: Continuous telemetry and daily 12-lead ECGs.
\end{enumerate}

\end{multi-turn-conversation}

\begin{multi-turn-conversation}
\textbf{User:} I need a rapid interpretation of this ECG for an emergency department handoff. Please provide a numbered reasoning walkthrough covering rate, rhythm, axis, and intervals. Use a focused findings and impression format. \\

\textbf{Assistant:} Reasoning: 
\begin{enumerate}
    \item Rate: The individual R-R intervals are 762, 770, 590, 863, 773, 762, 770, 754, and 773 ms. The average R-R interval is 757 ms. Calculating the heart rate: 60,000 / 757 = 79 beats per minute.
    \item Rhythm: The rhythm is sinus with a single Premature Atrial Contraction (PAC) identified by an early beat at 590 ms and a compensatory pause of 863 ms.
    \item Axis/Morphology: The QRS shows a deep negative deflection in V1 (amplitude -4.85 mV) and a prominent positive R-wave in V6 (amplitude 4.61 mV).
    \item Intervals: The PR interval is 193 ms. The QRS duration is markedly prolonged at 160 ms (exceeding the 120 ms threshold for a complete block).
    \item Conclusion: The combination of a wide QRS and the specific morphology in V1 and V6 indicates a delay in left ventricular depolarization.
\end{enumerate}
Findings:
\begin{itemize}
    \item Heart Rate: 79 bpm
    \item Rhythm: Sinus with 1 PAC
    \item PR Interval: 193 ms
    \item QRS Duration: 160 ms
    \item Lead V1 QRS Amplitude: -4.85 mV
    \item Lead V6 QRS Amplitude: 4.61 mV
\end{itemize}
Impression: Left Bundle Branch Block. \\

\textbf{User:} So the heart rate is about 80 beats per minute with one extra beat. Since the QRS is wide and the main deflection is pointing down in V1, this confirms a Right Bundle Branch Block, correct? \\

\textbf{Assistant:} Reasoning: In a Right Bundle Branch Block (RBBB), lead V1 typically shows a terminal upward deflection (RSR' pattern) because the final electrical forces move toward the right ventricle. In this ECG, lead V1 shows a deep, wide S-wave with a negative amplitude of -4.85 mV. The broad, positive R-wave in V6 (4.61 mV) and the wide QRS (160 ms) indicate that the impulse is traveling normally through the right bundle but is delayed reaching the left ventricle, which characterizes a left-sided conduction delay rather than a right-sided one.

Findings:
\begin{itemize}
    \item QRS morphology: Deep S-wave in V1, broad R-wave in V6
    \item QRS duration: 160 ms
\end{itemize}
Impression: Left Bundle Branch Block (Correction: This is not RBBB). \\

\textbf{User:} I understand now; the deep downward wave in V1 actually points to a left-sided delay. Can you provide the specific measurements for the P-wave and confirm the final PR interval for the clinical report? \\

\textbf{Assistant:} Reasoning: The P-wave duration is measured globally at 164 ms with an amplitude of 0.08 mV. Specifically looking at Lead II, the P-wave duration is 113 ms with an amplitude of 0.12 mV. The PR interval is 193 ms, which is within the upper limits of normal (less than 200 ms). These findings, alongside the 160 ms QRS duration, confirm the diagnosis of a significant ventricular conduction delay without a primary AV block.

Findings:
\begin{itemize}
    \item Global P-duration: 164 ms
    \item Lead II P-duration: 113 ms
    \item PR interval: 193 ms
    \item QRS duration: 160 ms
\end{itemize}

Impression: Left Bundle Branch Block.
      
\end{multi-turn-conversation}

\subsection{Stage 5} \label{appendix-stage5}

The Stage 5 dataset contains 17K samples, with ECGs sourced from the Icentia11k dataset.
Like Stage 4, we use ECGDeli-computed statistics to generate conversations with \texttt{gemini-3-flash-preview}.
However, in Stage 5 we focus on single-ECG, single-turn conversations (one question and one response).
We first compile a rubric consisting of the most clinically relevant features for forecasting AFIB and AFL, including PR interval prolongation \citep{pr-interval-components-afib-risk,long-term-pr-interval-outcomes}, heightened premature atrial contractions \citep{frequent-pac}, RR irregularity \citep{rr-irregularity-afib}, and increased standard deviation of normal RR interval (SDNN) \citep{hrv-afib}.
We verify these features with cardiologists to ensure clinical relevance.
We use this rubric to craft a Gemini prompt and select statistics to pass along with the prompt.
We provide an abridged system prompt for generating Stage 5 data below.

\begin{bigllmprompt}
You are an expert cardiologist specializing in arrhythmia risk prediction. Using ONLY the ECG statistics provided, produce evidence-based clinical reasoning to forecast whether an AFib or AFlutter event will occur within the next \{horizon\_seconds\} seconds.

Interpret the values as if you personally reviewed the ECG and measured them. Do not assume patient history, do not diagnose the current rhythm, and do not reference software, models, or automated interpretation.

\paragraph{Task}
Given ECG-derived metrics (including beat-level sequences such as RR, PR, QRS, and ectopy markers), produce:
\begin{enumerate}
    \item A concise, evidence-grounded clinical reasoning narrative describing near-term rhythm stability.
    \item A forecast of AFib/AFlutter occurrence within \{horizon\_seconds\} seconds.
\end{enumerate}

\paragraph{Core rules}
\begin{itemize}
    \item Use ONLY the provided statistics; do not invent values or features.
    \item Cite AT MOST 3--4 relevant metrics. Discuss evidence BEFORE forming any opinion.
    \item Ground every claim in specific metrics; reference individual beats explicitly when applicable (1-based).
    \item Numeric fidelity is mandatory. Show math for any derived values.
    \item Do not analyze metrics that contradict the ground-truth event.
    \item No filler, hedging clichés, patient history, or diagnostic statements.
\end{itemize}

\paragraph{Reasoning objectives (in order)}
\begin{enumerate}
    \item Evidence extraction: Identify metrics relevant to short-horizon AFib/AFlutter risk (e.g., RR variability, ectopy, atrial surrogates, conduction instability).
    \item Beat-referenced reasoning: Tie claims to specific beats when possible (e.g., “short RR at beat 7”).
    \item Mechanistic interpretation: Explain how observed findings plausibly precede AFib/AFlutter, using only supported relationships.
    \item Forecast formation: Provide a final forecast only after reasoning.
\end{enumerate}

\paragraph{Output requirements}
\begin{itemize}
    \item Return ONLY a JSON list of chat turns using double quotes.
    \item Include reasoning before the forecast.
    \item End with exactly one of:
    \begin{itemize}
        \item \texttt{Forecast: NORM}
        \item \texttt{Forecast: ABNORMAL}
    \end{itemize}
    \item If ABNORMAL, state whether AFib or AFlutter is more likely, or explicitly say uncertain.
\end{itemize}

\paragraph{Clinical phrasing constraints}
\begin{itemize}
    \item Do NOT refer to inputs as arrays, indices, fields, JSON, or statistics.
    \item Describe missing data clinically (e.g., “no premature atrial beats identified”).
    \item If PAC count is zero or label is NORM, do NOT mention PACs.
    \item Beat references must remain explicit and 1-based (e.g., “beat 5”).
\end{itemize}

\paragraph{Ground truth inputs}
ECG STATISTICS: \{stats\_str\} \\
Ground-truth Event after \{horizon\_seconds\} seconds: \{forecast\_label\}

\textbf{Generate the response now:}
\end{bigllmprompt}

We list several examples of conversations produced in Stage 5 below.

\begin{multi-turn-conversation}
\textbf{Human:} Analyze the ECG signal and predict the cardiac rhythm for the next 60.0 seconds.

NORM: Normal ECG

ABNORM: Atrial Fibrillation or Atrial Flutter

Output one of: NORM or ABNORM. \\

\textbf{Assistant:} In reviewing the rhythm for short-term risk, I first note the significant irregularity indicated by a high RMSSD of 194 ms and an RR Interval IQR of 128.91 ms. Examining the beat-to-beat timing reveals a severe shortening of the cycle length at beat 21, where the RR interval drops to 348 ms, compared to the mean RR interval of 793 ms. Furthermore, there are six premature atrial contractions identified in the segment, specifically occurring at beat 22, beat 23, beat 27, beat 30, and beat 36. The presence of frequent atrial ectopy, combined with a P-wave duration of 160.16 ms, suggests a high degree of atrial conduction delay and electrical instability. This pattern of frequent premature triggers on a background of high autonomic variability is a common precursor for paroxysmal atrial fibrillation. Given the high RMSSD and frequent ectopy, a transition into atrial fibrillation is likely within the next minute. Forecast: ABNORMAL
\end{multi-turn-conversation}

\begin{multi-turn-conversation}
\textbf{Human:} Analyze the ECG signal and predict the cardiac rhythm for the next 300.0 seconds.

NORM: Normal ECG

ABNORM: Atrial Fibrillation or Atrial Flutter

Output one of: NORM or ABNORM. \\

\textbf{Assistant:} In assessing this patient's rhythm for potential near-term transition into an arrhythmia, I am evaluating the stability of conduction and heart rate variability over the next 300.0 seconds. First, I noticed that the mean RR interval of 1030 ms, which correlates to a heart rate of 57.4 bpm, demonstrates a stable baseline. This is evident in the beat-to-beat consistency, for example, beat 1 at 1066 ms and beat 2 at 1043 ms. Second, the PR interval remains consistent with a mean value of 156 ms, and specific measurements such as beat 1 and beat 4 showing 156 ms and 152 ms respectively, indicating well-maintained atrioventricular conduction without the variations often associated with atrial instability. Third, the RR interval IQR of 71.29 ms, combined with an estimated atrial rate of 58, suggests that while there is physiological variability, the rhythm lacks the chaotic irregularity or rapid atrial activity that would precede a paroxysm of atrial fibrillation. Clinically, the uniformity of the QRS duration at 105.47 ms further supports a stable ventricular response. Given the absence of progressive shortening in the RR intervals or erratic shifts in the PR timing, the risk of a conversion to a tachyarrhythmia in the coming minutes is low. Forecast: NORM
\end{multi-turn-conversation}


\end{document}